\pdfoutput=1

\documentclass[11pt]{article}

\usepackage[preprint]{coling}

\usepackage{times}
\usepackage{latexsym}

\usepackage[T1]{fontenc}

\usepackage[utf8]{inputenc}

\usepackage{microtype}

\usepackage{inconsolata}

\usepackage{graphicx}

\newcommand{\methodology}{\textsc{DefVerify}}
\usepackage{booktabs}
\usepackage{amssymb}
\usepackage{pifont}
\newcommand{\cmark}{\ding{51}}%
\newcommand{\xmark}{\ding{55}}%
\usepackage{array}
\usepackage{subcaption} 
\usepackage{dblfloatfix} 
\usepackage{enumitem} 

\definecolor{ttcolor}{RGB}{64, 114, 139}
\definecolor{dcolor}{RGB}{133, 55, 67}
\definecolor{ercolor}{RGB}{127, 176, 94}
\definecolor{incolor}{RGB}{210, 163, 114}
\definecolor{boxcolor}{rgb}{0.431, 0.451, 0.824}

\usepackage[breakable]{tcolorbox}

\tcbset{
    mybox/.style={
        colback=boxcolor!5!white,
        colframe=boxcolor,
        fonttitle=\bfseries,
        coltitle=white,
        boxrule=0.5mm,
        rounded corners, 
        breakable
    }
}

\title{\textsc{\methodology}: Do Hate Speech Models Reflect Their Dataset’s Definition?}

\author{
 \textbf{Urja Khurana$^{\includegraphics[width=0.02\textwidth]{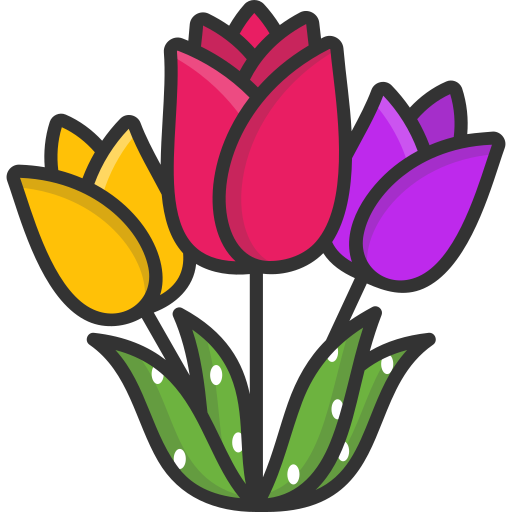}}$},
 \textbf{Eric Nalisnick$^{\includegraphics[width=0.02\textwidth]{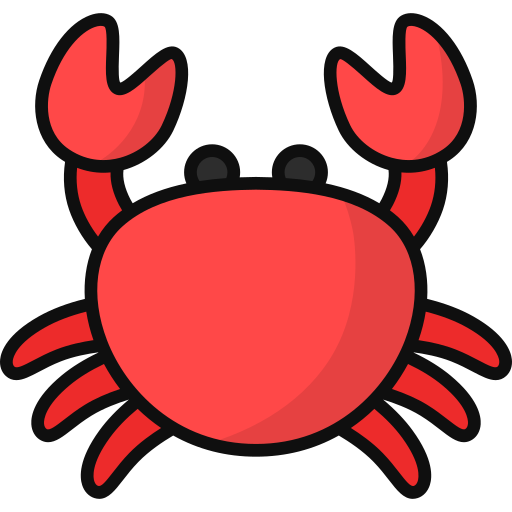}}$},
 \textbf{Antske Fokkens$^{\includegraphics[width=0.02\textwidth]{figures/emojis/tulips.png}}$}
\\
 {\includegraphics[width=0.02\textwidth]{figures/emojis/tulips.png}}Computational Linguistics and Text Mining Lab, Vrije Universiteit Amsterdam\\
 {\includegraphics[width=0.02\textwidth]{figures/emojis/crab.png}}Department of Computer Science, Johns Hopkins University
\\
   \texttt{u.khurana@vu.nl, nalisnick@jhu.edu, antske.fokkens@vu.nl}
}

\begin{document}
\maketitle

\begin{abstract}
    \textbf{Warning: Due to the nature of the topic, this paper contains offensive content.}  When building a predictive model, it is often difficult to ensure that application-specific requirements are encoded by the model that will eventually be deployed. Consider researchers working on hate speech detection. They will have an idea of what is considered hate speech, but building a model that reflects their view accurately requires preserving those ideals throughout the workflow of data set construction and model training. Complications such as sampling bias, annotation bias, and model misspecification almost always arise, possibly resulting in a gap between the application specification and the model's actual behavior upon deployment. To address this issue for hate speech detection, we propose \textbf{\methodology}: a 3-step procedure that (i) encodes a user-specified definition of hate speech, (ii) quantifies to what extent the model reflects the intended definition, and (iii) tries to identify the point of failure in the workflow.  We use \textbf{\methodology} to find gaps between definition and model behavior when applied to six popular hate speech benchmark datasets. 
\end{abstract}

\section{Introduction}
 Hate speech is a prevalent problem on social media but tackling it is not straightforward for numerous reasons. What constitutes hate speech varies by country and individual \citep{al-kuwatly-etal-2020-identifying, schmidt-wiegand-2017-survey}. There are many ways to address hate speech in a detection task, causing the definition of hate speech to change based on what the context demands. A law-based hate speech detection model in Belgium, e.g., may consider ``language'' as a protected group identity while other countries might not \citep{khurana-etal-2022-hate}. Researchers may decide to include stereotypes or, alternatively, restrict their definition to slurs. The groups that are considered targets can also differ.

 \begin{figure*}[h!]
    \centering
    \includegraphics[width=\linewidth]{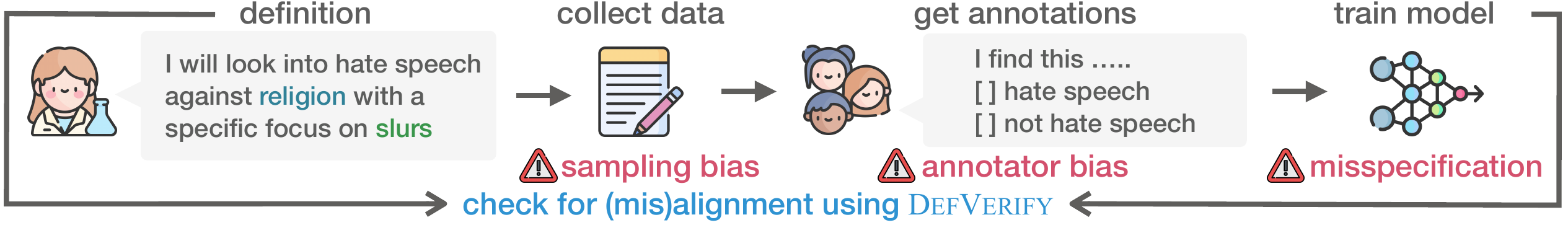}
    \caption{Potential failure points when creating a hate speech dataset. All biases can accumulate in the model.}
    \label{fig:problem-statement}
\end{figure*}

The type of hate speech that needs to be addressed affects the choice of data to train the detection models on. For instance, a dataset that only focuses on racism and sexism would not be suitable for an application that aims to capture hateful language against people with a disability. An important first step is thus making a \textit{definition} of hate speech for the project at hand, both for dataset creators and downstream users. When \textit{creating} a dataset, a creator needs to verify if the dataset is ultimately constructed according to the intended specifications for hate speech. For a user, \textit{finding} a dataset that aligns with the type of hate speech they want to address is important. 
Ideally, the dataset's hate speech definition\footnote{For brevity, we will refer to this as \textit{dataset's definition} throughout this paper.} would serve as an effective proxy for assessing whether the dataset is suitable to the project. However, the definition does not necessarily translate into model behavior. The potentially noisy process of creating datasets \citep{ross2016measuring, madukwe-etal-2020-data, fortuna-etal-2020-toxic, vidgen2020directions} can result in the dataset not covering and/or a model not learning the correct aspects for classification (see Figure~\ref{fig:problem-statement}). This has severe consequences when deploying such models in the real world, potentially inviting unexpected behavior when the model has to generalize to new data or domain. Carefully analyzing all steps of this process to ensure proper generalization is not always feasible.  \textit{How can one verify that a constructed dataset adheres to the intended task type of hate speech detection?}

\textbf{To investigate if a hate speech model behaves according to their dataset's definition, we propose a procedure to verify this: \methodology.} Our proposed methodology consists of three core steps when applied to a new hate speech dataset: (1) identifying which hate speech aspects, fueled by Hate Speech Criteria \citep{khurana-etal-2022-hate}, models trained on it should capture, (2) investigating to what extent it does by looking at HateCheck \citep{rottger2021hatecheck}, an English diagnostic evaluation set to uncover the strengths and weakness of a system, and cross-dataset performance, and (3) conducting an early-stage analysis of where the model fails based on the evaluation. To facilitate these steps, we build on HateCheck by (a) matching different aspects of definitions to each instance and (b) adding test cases that should be deemed offensive (or neutral) but not \textit{hate speech}. To demonstrate the utility of our approach, we apply it to \textit{six} different English hate speech datasets and examine to what extent these datasets follow their dataset's definition. We demonstrate gaps between dataset definitions and model capabilities for the datasets. 
Due to the task's subjective nature, the verification and evaluation of model capabilities are still open problems. Drifting away from a one-approach-fits-all approach, we are the first to investigate responsible model behavior through the lens of alignment between a dataset's hate speech definition and model capability. Our approach gives quick insights into what models can capture. The idea behind the approach can also be applied to other tasks. 

\section{Related Work}
\begin{figure*}[!t]
    \centering
    \includegraphics[width=\textwidth]{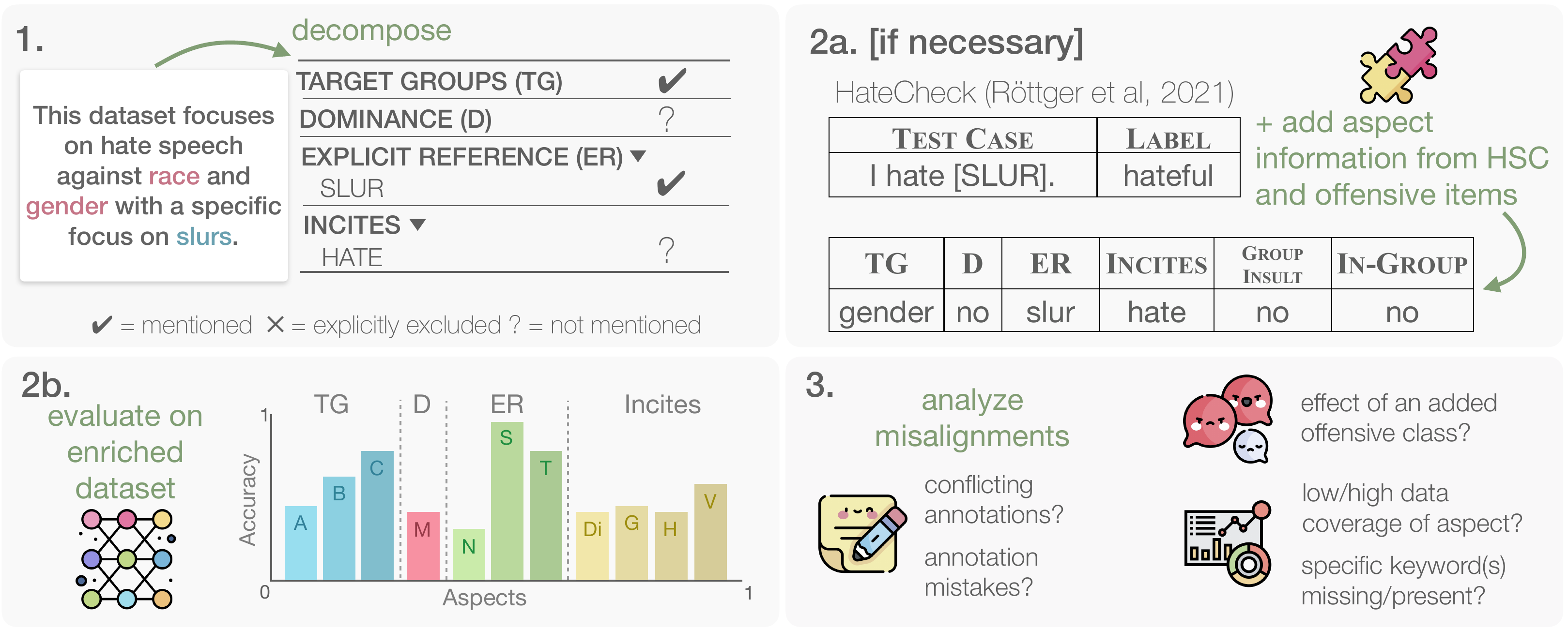}
    \caption{\textbf{\methodology}: Our proposed methodology to verify if model behavior is aligned with the dataset's intended definition. The figure depicts each individual step in the context of hate speech detection in our paper.}
    \label{fig:workflow}
\end{figure*}

\paragraph{Variations in hate speech definitions.}
The complexity of \textit{hate speech} is that there are various valid beliefs regarding what constitutes hate speech and what not \citep{rottger2022two}. Moreover, several aspects influence what hate speech is \citep{schmidt-wiegand-2017-survey}, and it also depends on the social context \citep{sap-etal-2019-risk}. 
Analyses of definitions on existing datasets confirm this complexity. \citet{fortuna-etal-2020-toxic} examine different hate speech datasets and find that even when datasets use very generic categories, their divergent definitions or inconsistent annotation can lead to distinct classifier performance. Datasets can also contain annotator-introduced variations. \citet{madukwe-etal-2020-data} highlight that varying definitions of hate speech can result in different labels to similar instances. Differences can furthermore stem from the way annotators interpret the guidelines \citep{vidgen2020directions}. \citet{awal2020analyzing} find inconsistencies in the labels for the Talat and Hovy, Davidson, and Founta datasets. \citet{isaksen-gamback-2020-using} confirm this for the Founta dataset. Similarly, \citet{van-aken-etal-2018-challenges} find doubtful labels in the Davidson dataset. \citet{ross2016measuring} conclude that for reliable annotations, better definitions and guidelines are needed for such a vague concept. Similarly, \citet{fortuna2021well} discuss the need for accurate and non-overlapping definitions. \citet{rottger2022two} point out how creators should think carefully about what kind of definition would suit their task: descriptive or prescriptive.

\paragraph{Generalization in hate speech detection.} 
The safety-critical nature of hate speech detection makes analyzing generalizability essential. Generalization behavior of hate speech detection models has been studied from different angles \citep{bourgeade-etal-2023-learn, yoder-etal-2022-hate, fortuna2021well, swamy-etal-2019-studying, antypas-camacho-collados-2023-robust, markov-etal-2021-exploring, markov-daelemans-2021-improving}. \citet{swamy-etal-2019-studying} find that balanced datasets lead to better generalization in cross-dataset studies. \citet{fortuna2021well} find models and the nature of categories to be decisive for generalization and intra-dataset performance to be indicative of generalization. 
\citet{yoder-etal-2022-hate} study variation of hate speech target identity, find that models struggle to generalize to other target identities. \citet{antypas-camacho-collados-2023-robust} find that combining different datasets is best for robustness for cross-dataset generalization. Closest to our work, \citet{bourgeade-etal-2023-learn} examine generalization between topic-generic and topic-specific datasets, topic referring to target group. Our investigation is more extensive, studying the relation with more aspects of hate speech definitions. Next to targets, we address dominance, explicit references to a group, and potential consequences. To our knowledge, we are the first to propose a methodology for such an investigation and conduct it.

\section{\methodology}
We propose \methodology\, a flexible framework that consists of the following steps: 
\begin{enumerate}
    \item Decompose definitions into hate speech aspects using Hate Speech Criteria (Section~\ref{subsec:step1})
    \item[2a.] Obtain (or create) a diagnostic set with aspect information (Section~\ref{subsec:step2})
    \item[2b.] Evaluate trained models according to alignment with its dataset's definition (Section~\ref{subsec:step3})
    \item[3.] Investigate the source of misalignment in the dataset (Section~\ref{subsec:step4})
\end{enumerate}

\subsection{Step 1: Decomposing Definitions}
\label{subsec:step1}
A typical hate speech dataset definition elaborates on what kind of hate speech its dataset intends to target. Hate Speech Criteria (HSC) \citep{khurana-etal-2022-hate}, a framework for creating definitions and annotation guidelines for hate speech tasks, identifies core aspects of hate speech that should be mentioned in a definition: either explicitly including or excluding the aspects. In practice, not many datasets specify all of them, either unintentionally or due to design choices. Mapping a definition to these aspects gives us a clear overview of what is intended to be included or excluded as hate speech and what remains underspecified. 

Our first step is manually \textit{decomposing definitions} (and annotation guidelines if publicly available) as provided by the creators of each dataset, using \textbf{HSC}.
HSC identifies the following \textit{four}\footnote{We drop \textit{perpetrator characteristics} as this information is not known for the diagnostic set nor the datasets we use.} relevant aspects: (1) target groups, (2) whether dominant groups can be targets, (3) types of explicit references to the group, and (4) the possible consequences of a hateful statement (e.g., inciting \textit{violence} or \textit{discrimination}). For instance, whether the phrase \textit{``white men are trash"} is seen as hate speech depends on aspect (2): \textit{can a historically dominant group be considered a target or not?} Step 1 in Figure~\ref{fig:workflow} illustrates this: \cmark~ for an aspect mentioned in the definition, \xmark~ if it is explicitly excluded, and ? if it is not mentioned. 

\subsection{Step 2a: Obtaining a Diagnostic Set}
\label{subsec:step2}
A diagnostic set provides instances representing specific phenomena. Such a set can reveal which aspects of hate speech are covered well by a model and verify if the model correctly excludes aspects considered out-of-scope, e.g.\ when it should ignore dominant groups as targets of hate. In \textit{our case}, we start from \textbf{HateCheck} \citep{rottger2021hatecheck}. This diagnostic set for hate speech detection already provides a rich set of clear and straightforward examples that a model for this task should correctly recognize. It consists of 2,968 samples, of which 60.7\% are hateful and 39.3\% are not. HateCheck follows its own definition which is not consistent with all definitions in our set. We enrich HateCheck samples with labels specifying aspects of HSC, so we can easily adjust the labels of the diagnostic set to e.g.\ include or exclude a specific target group or form of hate speech. As several hate speech datasets contain an additional \textit{offensive} class, we also extend HateCheck with \textit{offensive} samples. We make all our code and enriched version of HateCheck available.\footnote{\url{https://github.com/urjakh/defverify/}} This step is therefore only necessary when new phenomena are investigated in future work, e.g.\ perpetrator characteristics. We specify the additions we made in this step below. 

\paragraph{Adding aspect information.}
For each aspect, we indicate the possible sub-aspects/categories that we label with. See Step 2a in Figure~\ref{fig:workflow} for an example.
\noindent \textbf{\textsc{Target Group:}} Target groups in HateCheck: \textit{gender, sexual orientation, race, religion, nationality,} and \textit{disability}. \\
\noindent \textbf{\textsc{Dominance:}} yes or no, reflecting if it is a dominant group or not. All groups in the original HateCheck are non-dominant groups. We hence add two dominant groups: \textit{men} and \textit{white people}. It is critical to rigorously evaluate for these dominant groups, even when they are not considered targets of hate speech, to ensure that the model does not \textit{erroneously} identify them as targets of hate. Please note that we do not prescribe, we leave it up to the researcher to assess how they want to treat dominant groups.
\noindent \textbf{\textsc{Explicit Reference:}} 
how is the group mentioned: group characteristic, stereotype, or slur. 
\underline{Group Characteristic}: assigned at the mention of the name. Other types of group characteristics are not present in the data. 
\underline{Stereotype}: if something reads as a stereotype, it is marked as such, e.g.\ "typical of <GROUP>". We also factor in real-world stereotypes, e.g.\ how society views women in leadership positions. If an example contains a general stereotype (not group-specific), it is marked as only \textit{stereotype}, \textit{group characteristic} being absent, to prevent a high correlation between the two. 
\underline{Slur}: mostly marked in HateCheck based on the type of capability. Some slurs are also marked as a stereotype, such as \textit{c*m*lf*ck*r}. \\
\noindent \textbf{\textsc{Incites:}}
\underline{Hate}: Most cases incite hate due to the nature of the diagnostic set.
\underline{Violence}: Marked when violence is a potential consequence, both explicit and implied. This also implies hate but for distinction we only consider these to be violent. 
\underline{Discrimination}: Marked if the statement leads to exclusion. Cases that incite both violence and hate also incite discrimination but are not marked as such for distinction.
\noindent \textbf{\textsc{Group Insult:}} yes or no. Marked \textit{yes} when a negative attribute is assigned to a group. If someone expresses their own negative sentiment with a group characteristic, e.g.\ \textit{I dislike <GROUP>}, we do not consider it a group insult. \\
\noindent \textbf{\textsc{In Group:}} yes or no. Only \textit{yes} for reclaimed slurs present as we assume no exceptional context.

\paragraph{Adding offensive cases.} Hate speech datasets can also come with a third class; \textit{offensive}. This clashes with HateCheck which only contains \textit{hateful} and \textit{non-hateful} classes. Upon analyzing definitions and samples for \textit{offensiveness} of datasets used in this paper, this class mostly captures vulgar language in general (e.g.\ abusive words), violence toward individuals, and individual insults without any explicit link to group identity. We therefore treat two HateCheck functionalities as \textit{offensive}: \texttt{profanity\_nh} and \texttt{target\_indiv\_nh} (n=$165$). We also add $120$ instances with vulgar or abusive words in offensive and non-offensive (+$9$ cases) contexts, violent threats, and individual insults. We use the \textit{ableism} class from \citet{manerba-tonelli-2021-fine} as inspiration for individual insult and threatening, where originally all samples included a form of slur that we generalize to "you".

\subsection{Step 2b: Expectation vs. Reality}
\label{subsec:step3}
We now train a model on the dataset we are investigating. The model is then evaluated on the diagnostic set from Step 2a. We measure the model's accuracy on instances belonging to different sub-aspects. We can compare the model's performance to our expectations based on the definition and determine if the model's behavior reflects what was intended. In our case, HateCheck consists of very obvious and straightforward samples that we would expect a model to classify correctly. Hence, we expect performance to be quite good (e.g.\ at least ~$80$\% accuracy) on the desired aspects. 

\subsection{Step 3: Initial Root Cause Analysis}
\label{subsec:step4}
If the model performs unexpectedly on an aspect, we revisit the training data. Large datasets do curb the potential of conducting an extensive investigation for the failure. As a starting point, we use keywords related to the failing aspect to manually examine dataset coverage or inspect annotation consistency for similar training samples. 

\section{Retrospective Analysis} 
We now show the utility of \methodology\ and apply it to six widely used hate speech detection datasets. 

\subsection{Datasets}
\label{subsec:datasets}
We consider the following widely used hate speech datasets: \textbf{\textit{TalatHovy}} \citep{waseem2016hatefulTalat} (\textbf{TalatHovy}), \textbf{\textit{Davidson}} \citep{davidson2017automated}, \textit{Measuring Hate Speech Corpus} \citep{kennedy2020constructing} (\textbf{MHSC}), \textit{Dynamically Generated Hate Speech Dataset} \citep{vidgen2021learning} (\textbf{DGHS}), \textit{HateXplain} \citep{mathew2021hatexplain} (\textbf{HX}), and \textbf{\textit{Founta}} \citep{founta2018large}. The datasets are chosen based on their wide usage and diversity in hate speech definition. Note that \textbf{Davidson} and \textbf{HX} datasets have an \textit{offensive} class and \textbf{Founta} an \textit{abusive} class.

\subsection{Experimental Setup}
We fine-tune \texttt{fBERT} \citep{sarkar-etal-2021-fbert-neural} and \texttt{RoBERTa} \citep{liu2019roberta} on the six datasets, both shown to be competitive models for hate speech detection with different strengths in \citet{bourgeade-etal-2023-learn}. For each dataset, we train $5$ random seeds. Model selection is based on the validation macro F1. See Appendix~\ref{app:technical-setup} for technical details. For consistency with the original definitions, we keep the original labels of the datasets.\footnote{The class distribution of the datasets can be found in Table~\ref{tab:reverse-engineering}. For the HateXplain dataset, \textit{undecided} is excluded.} Hence, we preserve the \textit{offensive/abusive} class for the datasets that contain it. This enables learning the nuances, especially when dealing with borderline cases. Such cases may not be \textit{hate speech} per se, but still contain upsetting content that warrants flagging, e.g.\ for a law-based context. 
We evaluate with HateCheck, using accuracy (as in the original paper), precision, and recall. Due to the nature of the investigation, we filter out capabilities that test robustness toward spelling mistakes. We further do a separate analysis for dominant groups (which we discuss in Section~\ref{sub:reality-check}) and based on the definition, we mark these as \textit{hate speech} targets or not. \textbf{TalatHovy} dataset uses two specific sub-types of hate speech (\textit{sexism} and \textit{racism}). Thus, we consider instances where \textit{women} or \textit{trans people} are targeted as \textit{sexism} and \textit{black people, Muslims,} and \textit{immigrants} as \textit{racism}. All other instances are labeled non-racist or non-sexist.

\subsection{Applying \methodology}

\subsubsection{Decomposing Definitions}
\begin{figure}
    \centering
    \includegraphics[width=\linewidth]{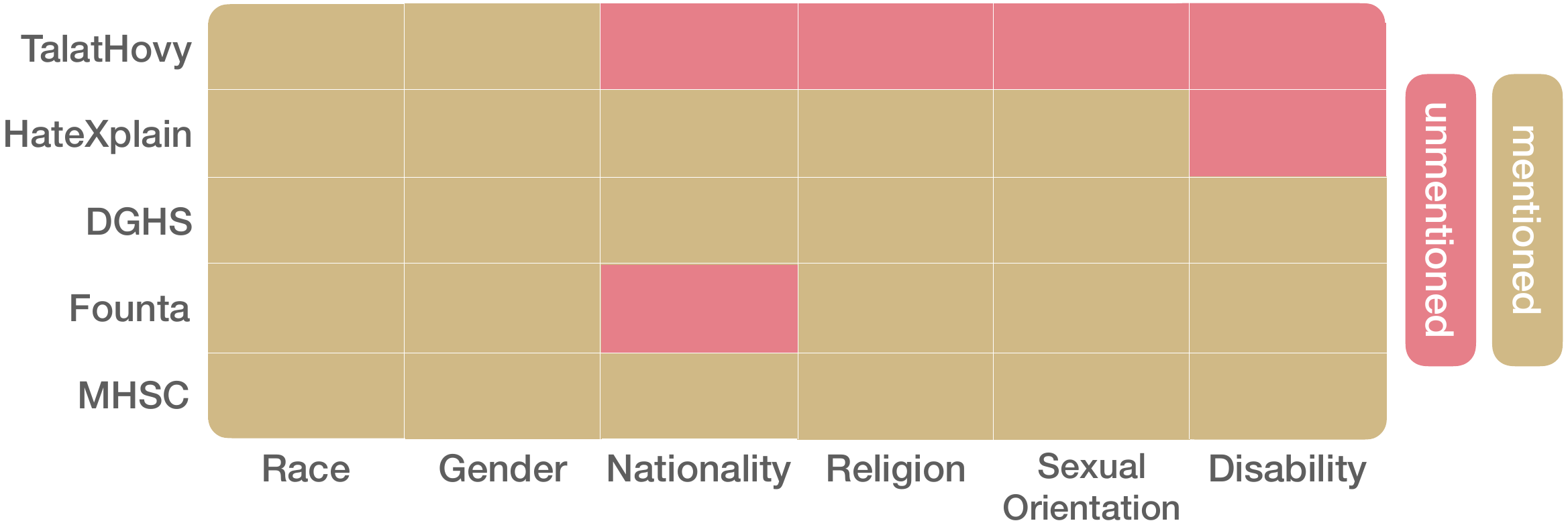}
    \caption{Overview of which target groups are mentioned in the definition for each dataset.}
    \label{fig:target-groups}
\end{figure}

For each dataset, we take the definition (and annotation guidelines for \textbf{DGHS}) from the original paper and provide the decomposed aspects in Table~\ref{tab:reverse-engineering} and covered target groups in Figure~\ref{fig:target-groups}.\footnote{Since \textbf{Davidson} does not mention any target groups in their definition, we leave it out of the plot.} We describe this process for each dataset, including the used definitions, in detail in Appendix \ref{app:process-reverse-engineering}. 

\begin{table*}[!ht]
    \centering
    \scalebox{0.82}{
    \begin{tabular}{cc>{\centering}p{7cm}cccccccccc}
        \toprule
        & \textbf{\textsc{Size}} & \textbf{\textsc{Label Composition}}  &  \textbf{TG} & \textbf{Do} & \textbf{ID} & \textbf{IV} & \textbf{IH} & \textbf{GI} & \textbf{St} & \textbf{GC} & \textbf{Sl} \\
        \midrule 
        DGHS & 41,255 & 54\% HS, 46\% not & \cmark & \xmark & ? & \cmark & \cmark & \cmark & \cmark & \cmark & \cmark \\
        \midrule
        TalatHovy & 16,914 & 20\% sexist, 11.7\% racist, 68\% neither & \cmark & \xmark & ? & \cmark & \cmark & \cmark & \cmark & \cmark & \cmark \\
        \midrule
        MHSC & 39,565 & 26.2\% HS, 78.8\% not & \cmark & \cmark &  \cmark & \cmark & \cmark & ? & ? & \cmark & ? \\
        \midrule
        Davidson & 24,802 & 5.8\% HS, 77.4\% offensive, 16.8\% neither & ? & ? & ? & \cmark & \cmark & \cmark & ? & ? & ? & \\
        \midrule
        Founta & 99,996 & 5.0\% HS, 27.2\% abusive, 14.0\% spam, 53.8\% normal & \cmark & ? & ? & ? & \cmark & \cmark & ? & ? & ? \\ 
        \midrule
        HX & 20,148 & 29.5\% HS, 27.2\% offensive, 38.8\% normal, 4.5\% undecided & \cmark & \cmark & ? & ? & ? & ? & ? & ? & ? \\ 
        \bottomrule
    \end{tabular}}
    \caption{    
    \textit{Decomposing} the datasets by mapping their respective definitions to the different aspects from \textbf{HSC}. The results are accompanied with information about the dataset size and label distribution (where \textbf{HS} stands for hate speech). \cmark~indicates that the aspect is mentioned in the definition and is considered, \xmark~indicates that the aspect is explicitly not considered for the dataset, and \textbf{?} means that it is unmentioned and hence we do not know what to expect. \textbf{TG}: target groups, \textbf{Do}: dominant groups, \textbf{ID}: incitement of discrimination, \textbf{IV}: incitement of violence, \textbf{IH}: incitement of hate, \textbf{GI}: group insult, \textbf{St}: stereotype, \textbf{GC}: group characteristics, \textbf{Sl}: slur.  
    }
    \label{tab:reverse-engineering}
\end{table*}

\paragraph{Included aspects in dataset definitions.} We see that the \textbf{DGHS, TalatHovy, } and \textbf{MHSC} datasets include the most aspects in their definition. Only one or two definitions specify whether they consider solely non-dominant groups or also include dominant groups as potential hate speech targets. The definition of \textbf{HX} is limited to a specification of the target groups. We can infer from this that dominant groups are included.

\paragraph{Expected model behavior} Models trained on a respective dataset should have high performance identifying aspects indicated with a \cmark~in Table~\ref{tab:reverse-engineering} as hate speech and aspects marked with \xmark~as non-hate speech. We do not construct expectations for aspects marked with a \textbf{?}~as its absence from the definition can introduce more space for annotator subjectivity or hint at limited to no data coverage.

We make certain assumptions for our expectations when decomposing the definitions of the datasets. Though most datasets do not clearly define that they require an \textit{explicit reference} to the group being attacked, we do assume that this is intended to distinguish between hate speech and other forms of toxic language. We thus expect good performance on \textit{group characteristics}, the aspect which also captures references to a group by their name, for all models. Additionally, when a hateful utterance can lead to discrimination, we automatically assume this to be a group insult as well. 

Most expectations can be inferred directly from the table. For \textbf{TalatHovy}, we expect models to perform well on non-dominant groups based on their \textit{race} and \textit{gender}, as well as \textit{religion} and \textit{nationality} based on our observations in the dataset. Due to the focus on race and gender-based minorities, we do not expect the model to classify dominant groups from these categories (i.e.\ men and white people) or other target groups as hate speech. \textbf{Davidson} does not mention any target group or dominance. Thus, it is unclear on which target groups the model will do well or how it will respond to dominant groups. 

\subsubsection{Expectation vs. Reality}
\label{sub:reality-check}
We now evaluate the models of each dataset on our diagnostic evaluation set, the enriched version of HateCheck.\footnote{For brevity, we refer to a model's performance trained on a specific dataset as \textit{dataset's} performance.} Each dataset's validation and test set results can be found in Appendix~\ref{app:results-indiv-datasets}. We see a large drop in performance from the original test to diagnostic tests, indicating that the models learned the data, but not the intended aspects.

\begin{figure*}[h!]
    \centering
    \subfloat[Davidson dataset.]{
        \includegraphics[width=0.5\textwidth]{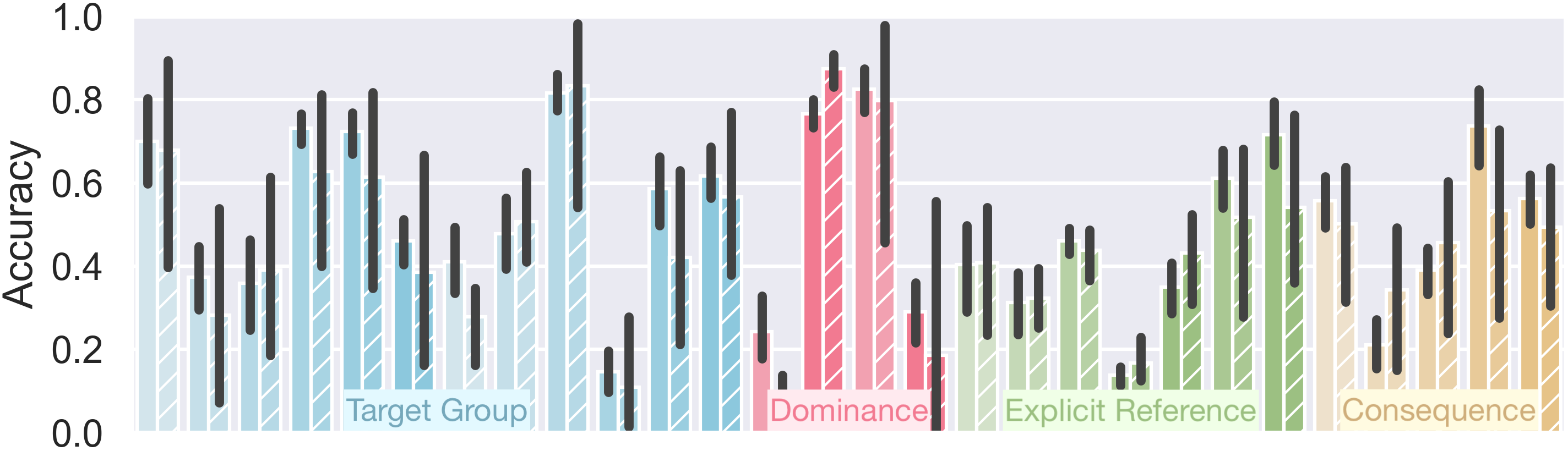}
        \label{fig:hc-davidson}
    } 
    \subfloat[Founta dataset.]{
        \includegraphics[trim=1.6cm 0 0 0, clip, width=0.47\textwidth]{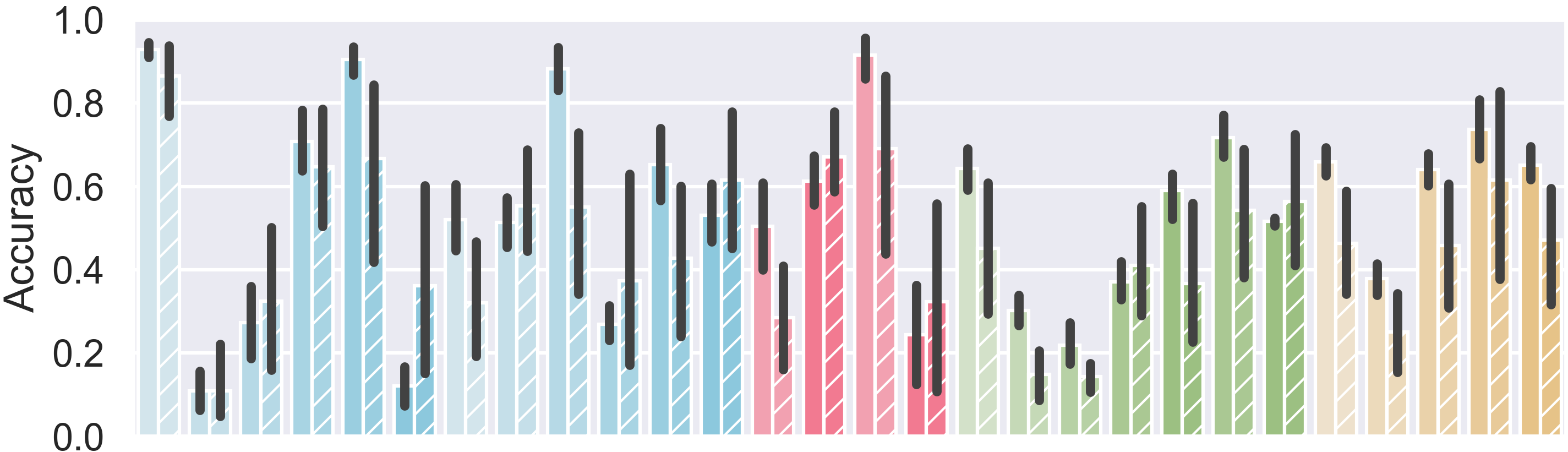}
        \label{fig:hc-founta}
    } \\
    \subfloat[MHSC dataset.]{
        \includegraphics[width=0.5\textwidth]{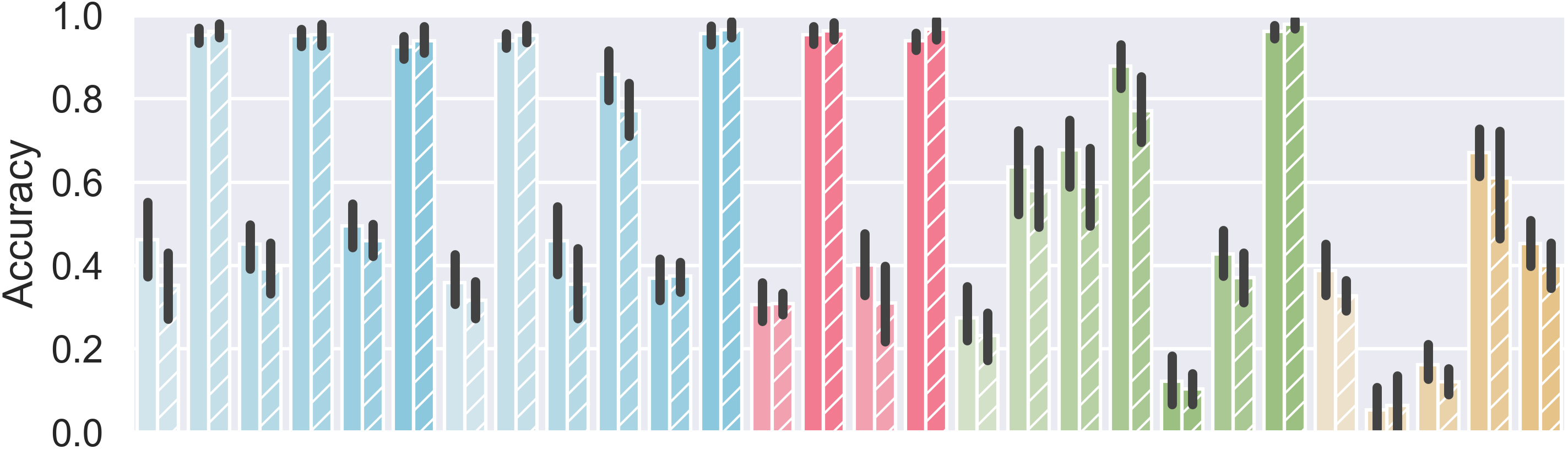}
        \label{fig:hc-kennedy}
    } 
    \subfloat[HX dataset.]{
        \includegraphics[trim=1.6cm 0 0 0, clip, width=0.47\textwidth]{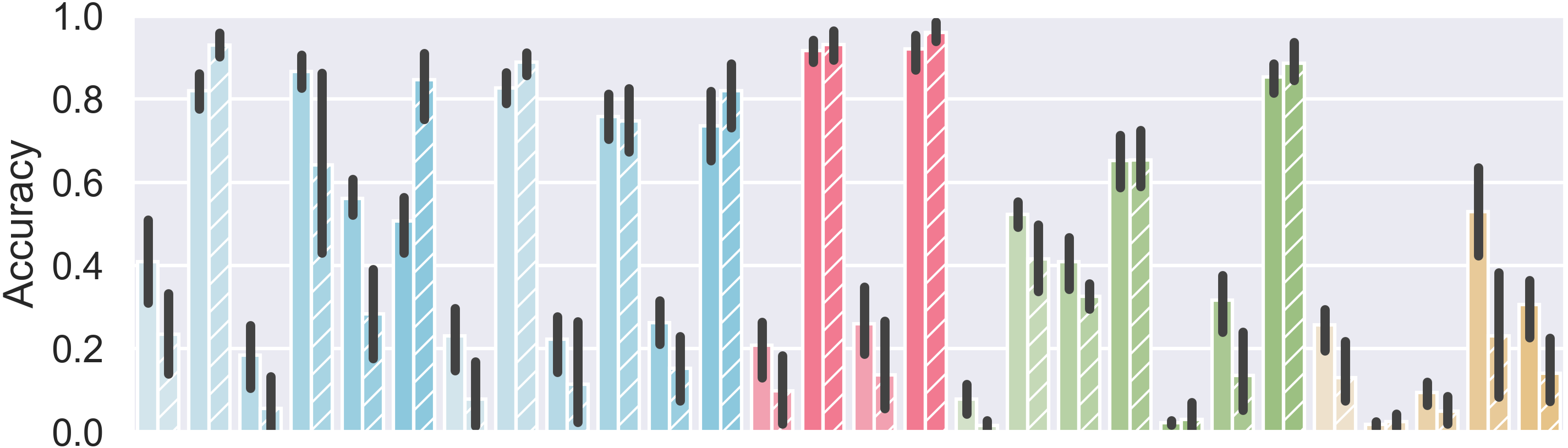}
        \label{fig:hc-mathew}
    } \\
    \subfloat[TalatHovy dataset.]{\vspace{0.25cm}
        \includegraphics[width=0.5\textwidth]{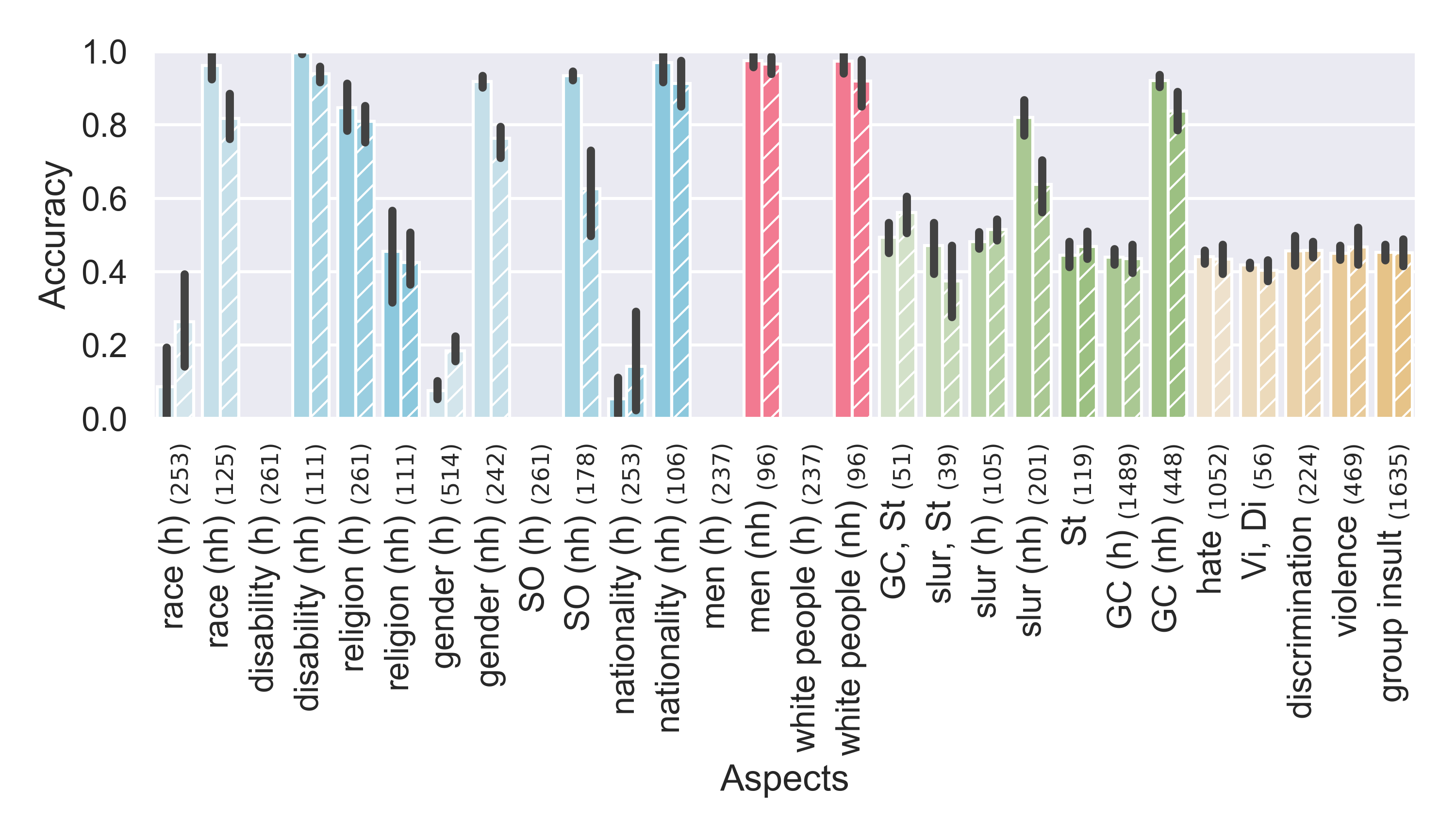}
        \label{fig:hc-talat-hovy}
    } 
    \subfloat[DGHS dataset.]{\vspace{0.25cm}
        \includegraphics[trim=1.6cm 0 0 0, clip, width=0.47\textwidth]{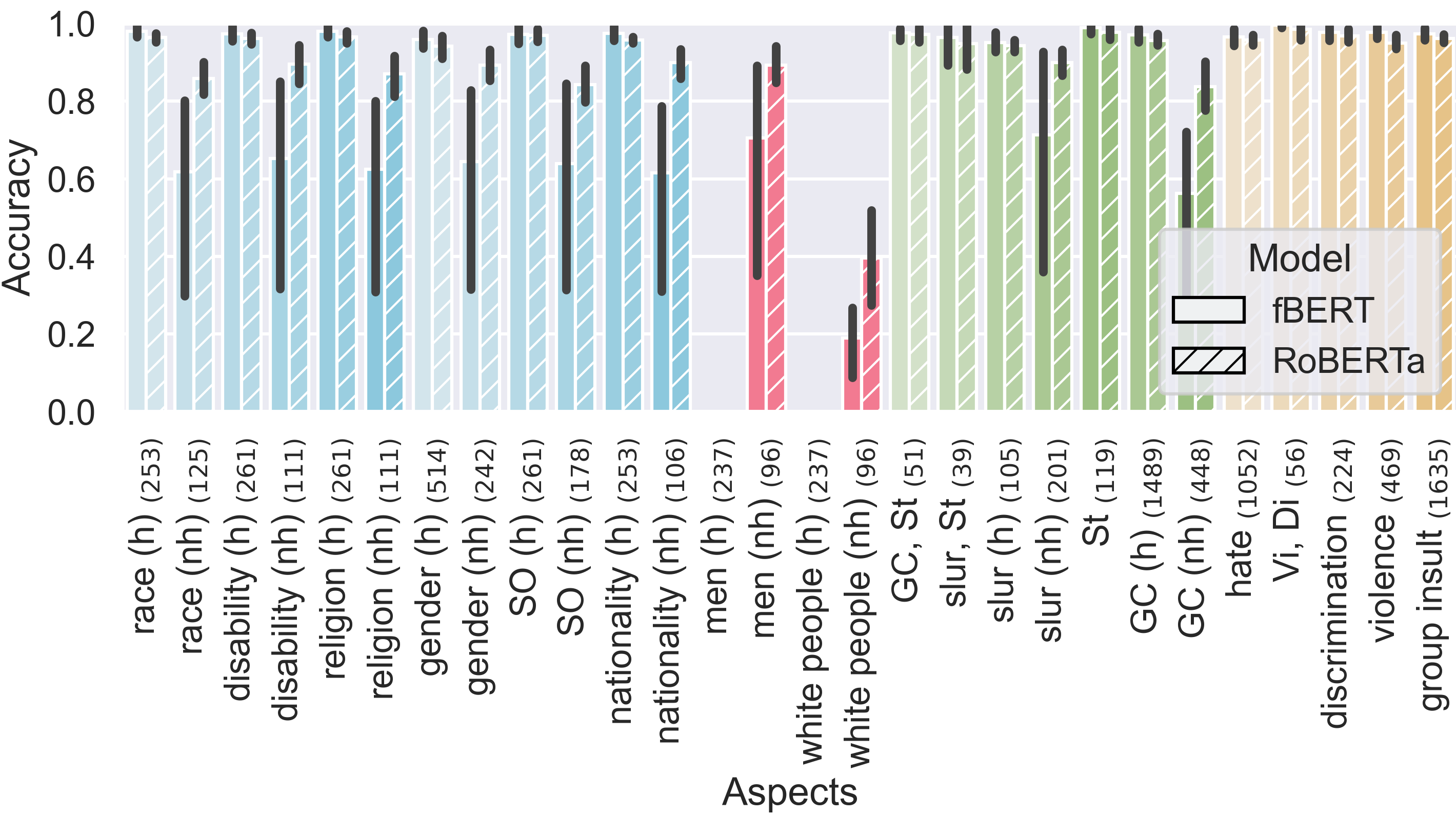}
        \label{fig:hc-vidgen}
    }
    \caption{Accuracy for the different aspects on HateCheck, with the number of samples for each aspect in brackets. We separate the accuracy for aspects with both hate (h) and non-hate (nh) samples. \textbf{SO}: sexual orientation, \textbf{GC}: group characteristics, \textbf{St}: stereotype, \textbf{Vi}: incitement of violence, \textbf{Di}: incitement of discrimination.}
    \label{fig:hc_properties}
\end{figure*}

\begin{figure*}[h!]
    \centering
    \subfloat[Binary Davidson dataset.]{\vspace{0.25cm}
        \includegraphics[width=0.32\textwidth]{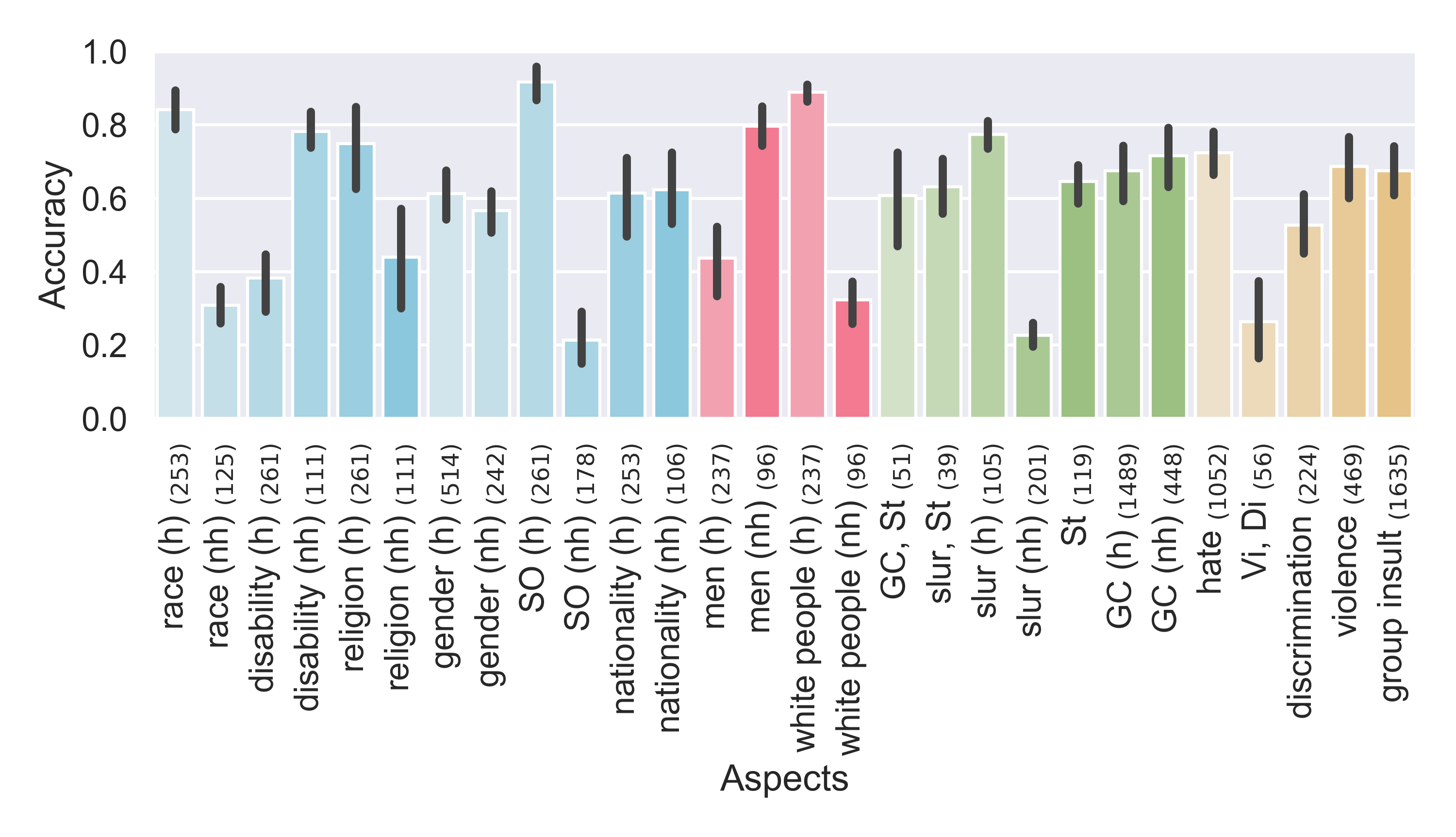}
        \label{fig:hc-davidson-binary}
    } 
    \subfloat[Binary Founta dataset.]{
        \includegraphics[trim=2.5cm 0 0 0, clip, width=0.32\textwidth]{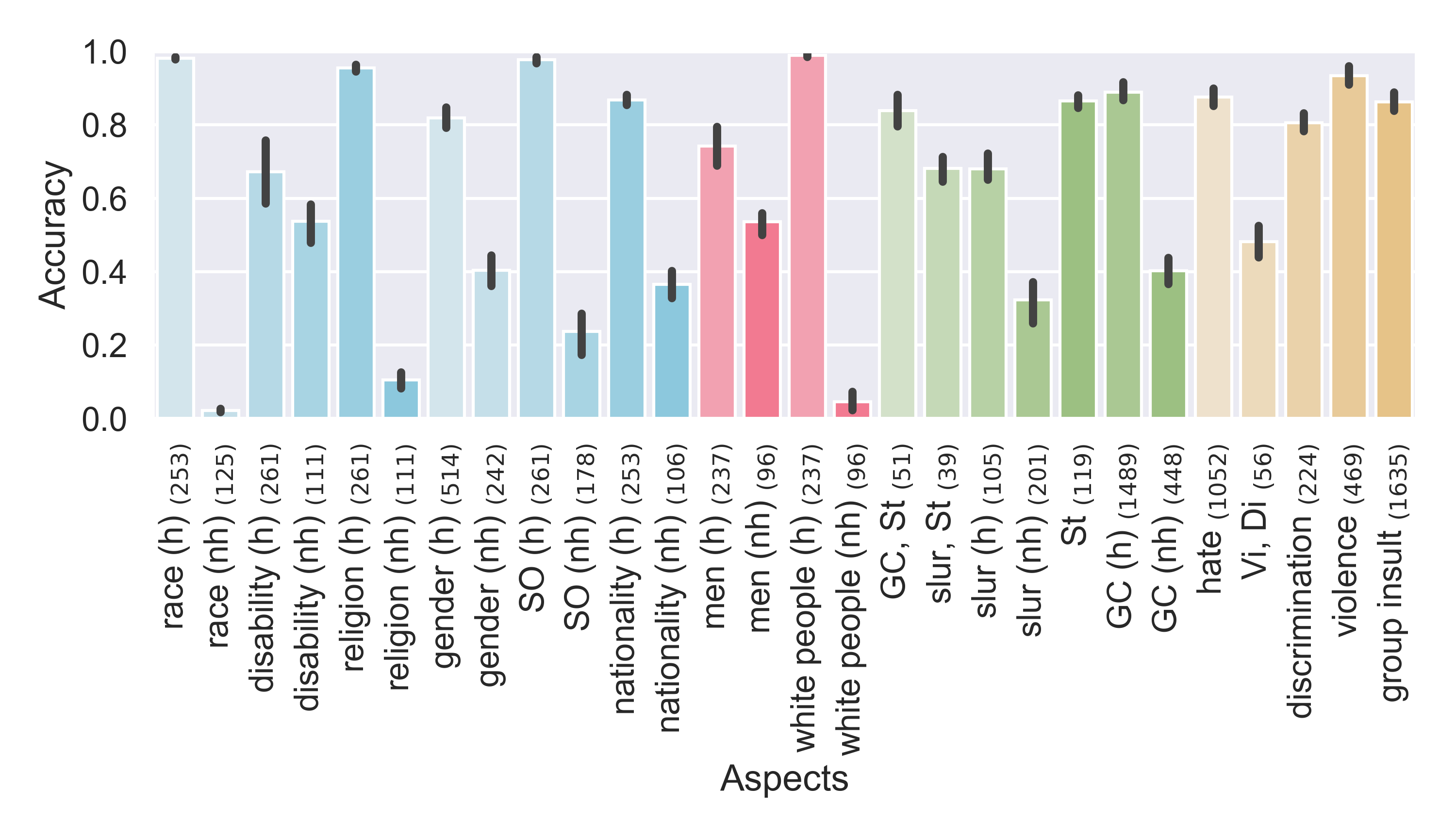}
        \label{fig:hc-founta-binary}
    } 
    \subfloat[Binary HX dataset.]{
        \includegraphics[trim=2.5cm 0 0 0, clip, width=0.32\textwidth]{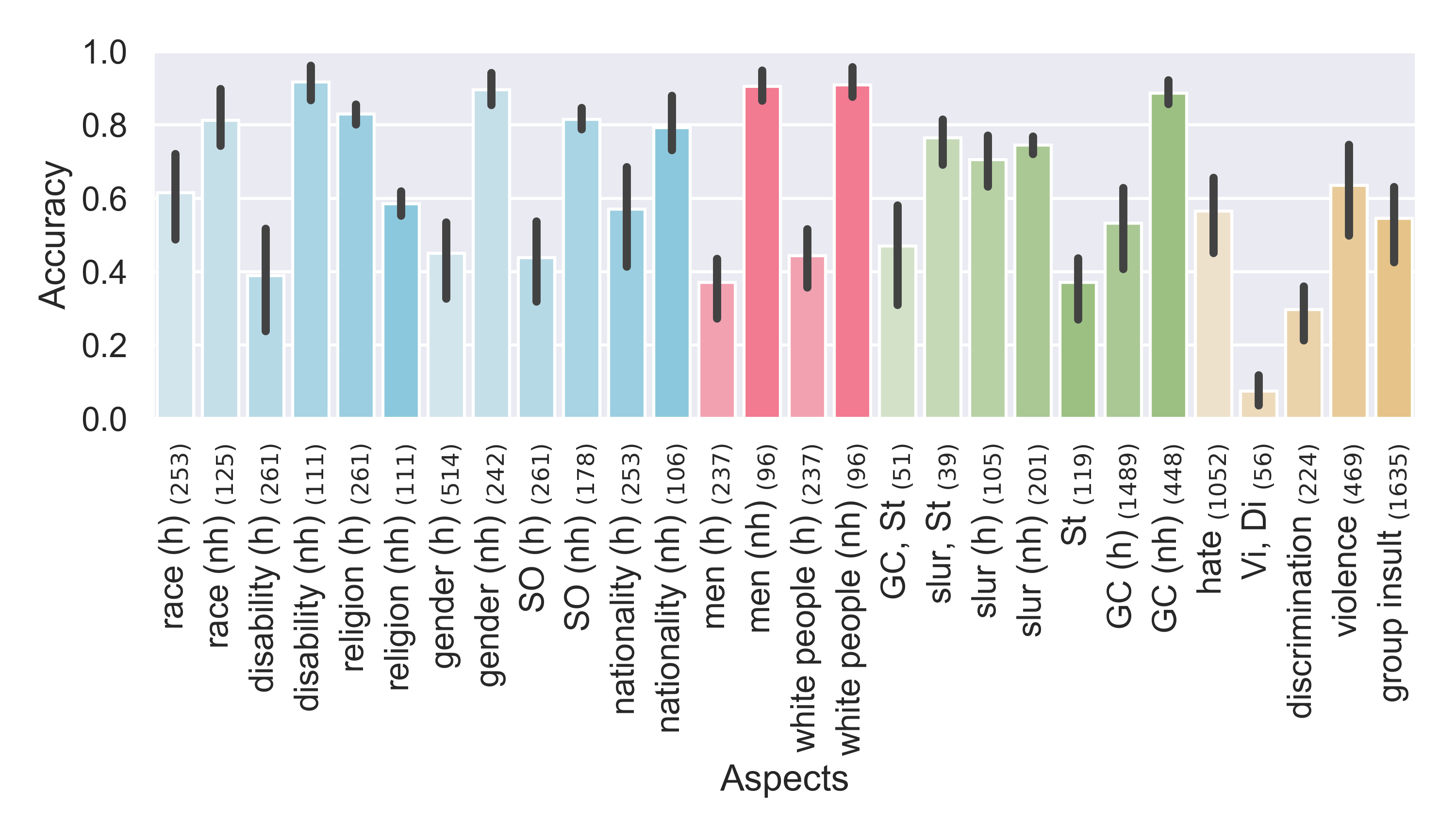}
        \label{fig:hc-mathew-binary}
    }
    \caption{Accuracy for the different aspects on HateCheck, when removing the \textit{offensive} classes from the \textbf{Davidson, Founta,} and \textbf{HX} datasets with \texttt{fBERT}.}
    \label{fig:hc_properties_binary}
\end{figure*}

\paragraph{Alignment of model behavior with definition.} 
Figure~\ref{fig:hc_properties} showcases the results on all the tested aspects in HateCheck for each dataset. For \textbf{Davidson}, we expected good performance for incitement of violence and hate and group insults. We only see this for violence, the other two achieving rather average performance. Similarly, with \textbf{Founta}, we see average performance for the aspects we expected to perform well: incitement of hate and group insult. Although \textit{gender}, \textit{sexual orientation}, and \textit{religion} get recognized as hateful, there are a lot of false positives. \textbf{MHSC} has low recall on identifying hate speech for target groups in general. We see the same for the dominant groups (explicitly covered). Low to average performance is achieved for all the other aspects it was expected to do well on except for \textit{violence}. \textbf{HX} underperforms severely; regardless of whether an aspect is mentioned in the definition or not, it predominantly results in false negatives. \textbf{TalatHovy }models correctly identify excluded target groups (dominant groups, disability, and sexual orientation) as true negatives. Race (mostly classified as \textit{sexist} by the model), gender, and nationality (latter both classified as \textit{neither} by the model) perform badly. Other mentioned aspects also underperform. \textbf{DGHS} performs well on most aspects, only misclassifying white people as false positives, while they were excluded as a target in the definition. \textbf{We find that many aspects, even when mentioned in the definition, do not get picked up by the model correctly, either overclassifying or failing to recognize hateful aspects.}

\paragraph{Removing \textit{offensive} when training.} Datasets with an extra \textit{offensive} class have a disadvantage when the evaluation set only considers the other two classes. We hence remove the \textit{offensive} class from \textbf{Davidson, Founta,\footnote{As the original proportion of hateful samples is only 1/10, we improve the imbalance by increasing this to 1/3}} and \textbf{HX}. We show the results in Figure~\ref{fig:hc_properties_binary}. For all three datasets, most aspects see an increase in accuracy, as expected. For \textbf{Davidson}, the impact is biggest for aspects related to \textit{explicit references} with good results on \textit{group characteristic} in general. When it comes to consequences, \textit{hate} and \textit{group insult} increase, just as \textbf{Founta}. Additionally, for \textbf{Founta}, we see that \textit{discrimination} and instances covering \textit{stereotype} or \textit{group characteristic} have an increased accuracy, while \textit{slurs} remains challenging. For \textbf{HX}, \textit{target groups} and \textit{dominance}'s accuracies increase in general, while all \textit{explicit references} increase. It still struggles with any \textit{discriminatory} aspects.

\begin{figure*}[h!]
    \centering
    \subfloat[Davidson.]{
        \includegraphics[width=0.2\textwidth]{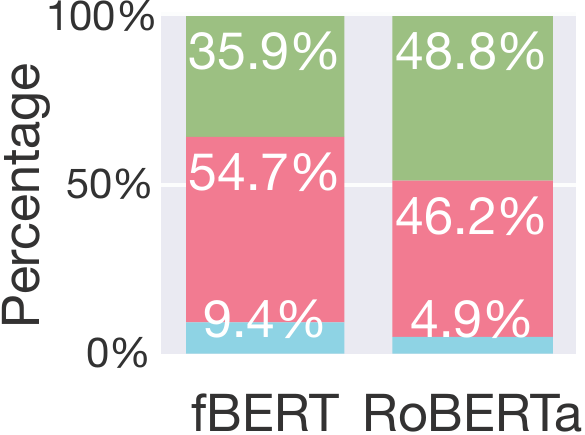}
        \label{fig:off-davidson}
    } 
    \subfloat[Founta.]{\hspace{-0.1cm}
        \includegraphics[width=0.145\textwidth]{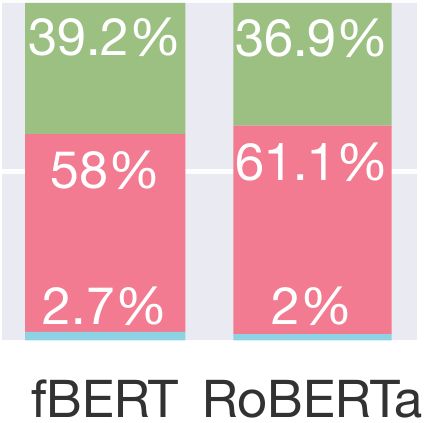}
        \label{fig:off-founta}
    } 
    \subfloat[MHSC.]{\hspace{-0.1cm}
        \includegraphics[width=0.145\textwidth]{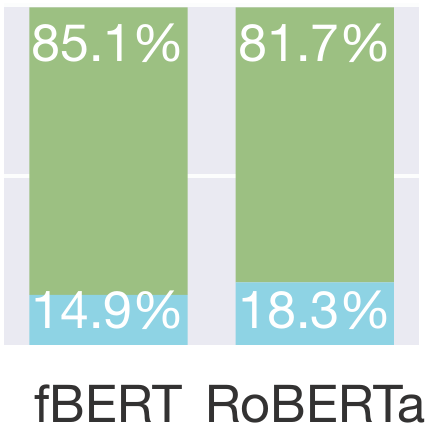}
        \label{fig:off-kennedy}
    } 
    \subfloat[HX.]{\hspace{-0.1cm}
        \includegraphics[width=0.145\textwidth]{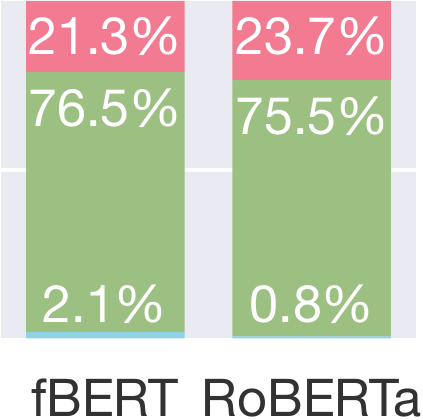}
        \label{fig:off-mathew}
    } 
    \subfloat[TalatHovy.]{\hspace{-0.1cm}
        \includegraphics[width=0.145\textwidth]{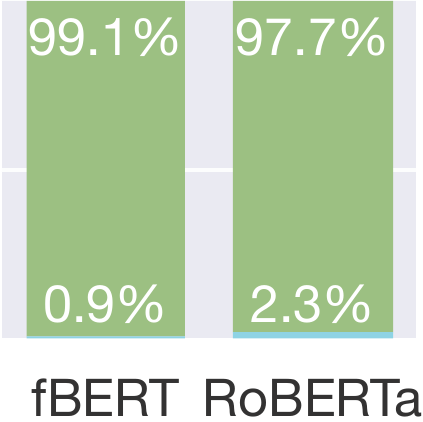}
        \label{fig:off-talat-hovy}
    } 
    \subfloat[DGHS.]{\hspace{-0.1cm}
        \includegraphics[width=0.145\textwidth]{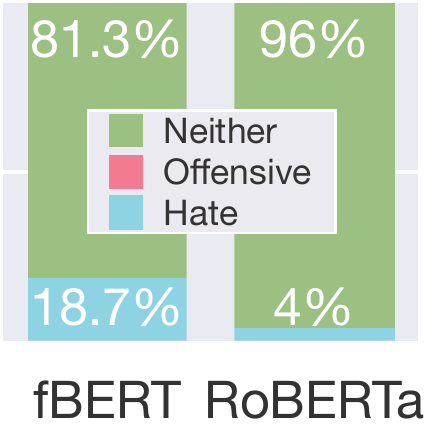}
        \label{fig:off-vidgen}
    }
    \caption{Percentage of predictions on $285$ offensive samples for each dataset.}
    \label{fig:offensivelist}
\end{figure*}

\paragraph{Performance on offensive items.} 
We evaluate performance on $265$ added offensive statements and show the results in Figure~\ref{fig:offensivelist}. In general, the two models perform similarly across most datasets. For \textbf{Davidson}, only roughly ~$50$\% is seen as offensive and for Founta this increases to ~$60\%$. This is against expectations. \textbf{HX} models  (Figure~\ref{fig:off-mathew}) primarily view the offensive statements as \textit{non-offensive}, exposing the lack of generalization to unseen \textit{offensive} data. \textbf{MHSC} (Figure~\ref{fig:off-kennedy}), \textbf{TalatHovy} (Figure~\ref{fig:off-talat-hovy}), and \textbf{DGHS} (Figure~\ref{fig:off-vidgen}) models correctly predict most samples to be \textit{non-hateful}.

\begin{figure}
    \centering
    \includegraphics[width=0.45\textwidth]{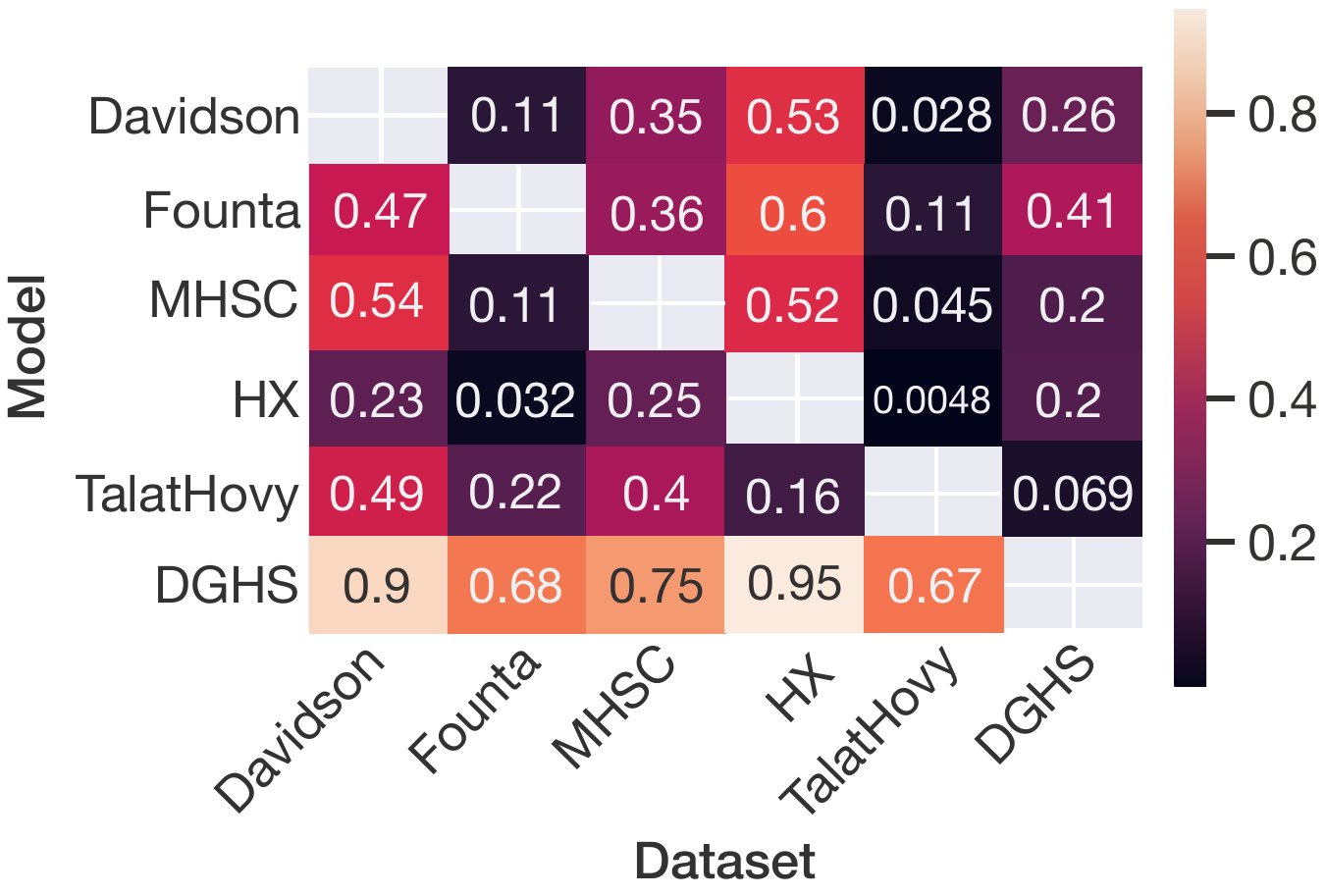}
    \caption{Accuracy (averaged over seeds) of identifying hate instances on cross-dataset test sets with \texttt{fBERT}.}
    \label{fig:cross-eval}
\end{figure}

\paragraph{Cross-dataset performance.} In Figure~\ref{fig:cross-eval}, we showcase the accuracies achieved when evaluating a model trained on dataset A on the test set of dataset B (where A$\neq$B). Due to different labels in different datasets, we only focus on how many of the hate speech instances of a dataset the model recognizes correctly as hate. For example, we expect \textbf{DGHS} to perform well on \textbf{TalatHovy} due to their similarity in terms of aspects. \textbf{We do not observe that datasets with similar definitions tend to yield better performance on each other's test sets compared to sets with dissimilar definitions.} Overall, we find that \textbf{DGHS} achieves the best performance on all the datasets, with great results on \textbf{Davidson} ($0.9$) and \textbf{HX} ($0.95$) particularly. The rest of the models from other datasets perform notably low. Datasets with an \textit{offensive} class do not stand out on each other performance-wise.

\subsection{Initial Root Cause Analysis}
Understanding why a model failed to capture an aspect it was expected to cover can lead to fixing those gaps.
We thus conduct an early-stage analysis for some of the observed misalignments between model behavior and the original dataset's definition. We do a simple keyword search in the dataset's training set, using terms relevant to the failing hate speech aspects. These terms primarily stem from HateCheck. We then investigate the label distribution for matching samples and data coverage. If those do not show useful insights, we look at the consistency of annotations. While this analysis does not give a holistic view of the reasons for misalignment, it is a starting point to get more targeted insights from what we are training our model on. In case of sufficient relevant training data points, this could hint at model underfitting. A lack thereof could mean adding more relevant data points. We propose more thorough ways to analyze misalignments in the future in Section~\ref{sec:discussion}.

\paragraph{TalatHovy performs badly on \textit{sexism} and \textit{racism}.} Since \textit{sexism} in \textbf{HateCheck} consists of \textit{trans people} and \textit{women}, we use those as keywords to find training samples. Only 0.8\% of the training data mentions \textit{trans}, with only $2$ labeled sexist. For \textit{women/woman} there are some mislabels, e.g. \textit{``A woman must be obedient"} as \textit{racism}. Several samples labeled as sexist contain the phrases \textit{``I'm not sexist"}, \textit{``Call me sexist"}, or \textit{``\#sexist"}. For \textit{racism}, we find the term \textit{black} in only $64$ samples with only $11$ marked as \textit{racist}. Variations of the n-word and f-word are only found in 2 samples.

\paragraph{DGHS recognizes white people as hate speech.} Many of the cases in the training set that contain \textit{white} express superiority, where $746$ samples are deemed hate and $394$ are not. In contrast, for \textit{man} which it does not recognize as hate speech, $202$ of $376$ samples are marked as not hate. \textit{Men}'s label distribution is around $50$\% for both classes, with a lot of intersectional cases when marked as hate.  

\paragraph{MHSC does not recognize men as hate speech.} In total there are $528$ samples in the training set that mention \textit{men}. Out of these, $120$ are hate, and $408$ are labeled as not. We also see some inconsistencies in annotation where \textit{``Wow men are f*ck*ng trash and disgusting"} is labeled as hate but two other examples calling \textit{men trash} as not hate.

\section{Discussion}
\label{sec:discussion}
\paragraph{Analyzing DGHS' higher performance.} We find that models trained on DGHS achieve the best results on HateCheck and cross-dataset evaluation experiments. The authors between HateCheck and DGHS overlap, which is addressed in the paper of DGHS: only 0.05\% of the dataset matches HateCheck, but they still call for caution. The cross-evaluation results hint that the good performance on HateCheck does not solely stem from that. DGHS comes with a beneficial dataset composition (large training set with balanced class distribution) and is synthetically generated using an adversarially contrastive approach. The latter can reduce false positives and negatives. We ran additional experiments adjusting the dataset to less ideal circumstances: a small training set and imbalanced class distribution. The exact results can be found in Appendix~\ref{app:vidgen-datacomp}, but performance only suffers significantly for the imbalanced cross-evaluation. 

\paragraph{Root Cause Analysis.} Our keyword-matching analysis is just the first step to locating the failure point for misaligned behavior. In the current scenario, it gives us a limited view of what is in the data, what the model relies on for predictions, and annotation bias at a large scale. More \textit{data} and \textit{model} interpretability techniques are needed, e.g.\ using influence functions \citep{koh2017understanding} to understand what evidence a model used in the case of misclassifications. Important insights can come from analyzing which samples in the data have high annotator variation and which are straightforward. This could give insights into which datasets are more prone to misclassify edge cases. Many datasets do not provide individual annotator annotations nor annotation guidelines, which are essential elements for a full analysis of data quality. \textbf{In future work}, we intend to aim for a more holistic view to attack the cause of misalignment from different angles with \methodology, disentangling the biases shown Figure~\ref{fig:problem-statement}.

\paragraph{The complexity of statements in the diagnostic dataset.} We use HateCheck due to its straightforward samples and the large number of different capabilities that a hate speech detection system should have. The simplicity of these statements in HateCheck is essential to establish whether the model can at least recognize the simplest forms of hate speech. However, this excludes more complex and implicit forms of hate speech, which are more representative of such language in the wild. It is even more so important that we should also be aware of how our model fares on those. \textbf{Future work} may focus on extending the diagnostic dataset to also cover rather complex hate speech samples.

\section{Conclusion}
We propose \methodology, a methodology to investigate if a hate speech dataset correctly captures the intentions of a dataset creator or user. We achieve this by (1) eliciting and formalizing the definitions underlying popular hate speech datasets through the lens of Hate Speech Criteria, (2a) enriching HateCheck with information based on this lens and with extra offensive statements, (2b) evaluating potential misalignments between dataset definitions and model capabilities on six widely-used hate speech datasets, and (3) an early-stage analysis to understand observed gaps. Our findings indicate that for most datasets, their respective trained models failed to correctly recognize aspects, even when they were explicitly mentioned in the dataset's definition.

\paragraph{Takeaways.} Our method provides an approach to pinpoint if a dataset provides the type of hate speech a creator intended to address. The gaps we found using \methodology\ showcase that evaluating model behavior is important when \textit{creating} datasets as well as \textit{deploying} models. Our method can also help users select a dataset for their specific purposes. Our methodology is thus more than just a tool to investigate definition alignment in hate speech; it supports more informed dataset creation and usage to help ensure more reliable models in safety-critical contexts. For more reliable models in the wild we need (1) clarity in terms of what aspects are explicitly included and excluded in the definition, (2) clearly reflected in annotation guidelines that (3) should be made publicly available (4) along with individual annotations, and (5) rigorously evaluated before deployment.  

\section*{Limitations}
While we aim to rigorously analyze model behavior and capture performance on several aspects of hate speech that have not been tested before, there are a few limitations in our work. 

\paragraph{Considering other relevant factors.} Our paper is a first step in unraveling the disparity between the intended definition and what is in the data. For our methodology, we have chosen a method that can be applied with information that is publicly available for all datasets and is relatively easy and quick to carry out (since it is unlikely that non-research users will carry out an extensive analysis of annotation and data itself). There are many factors between the starting point of a general definition and model behavior that influence the final model: the annotation guidelines, the data itself (size, properties, selection method), the quality and number of the annotators, and the models used for training, to name a few. Each of these can and should ideally be studied at their own account to gain full insight into the suitability of a dataset. Our approach merely provides a first step based on information and resources that are generally publicly available to potential users of the data, for off-the-shelf usage: the definition provided when the data was introduced, standard models, and the extended diagnostic test that we provide with this paper. For example, for our analysis, we would ideally have taken into account the annotation guidelines more but we could only do this for the \textbf{DGHS} dataset. We made use of information that was publicly available and most datasets do not provide annotation guidelines. This is why we focus on definitions provided in the papers introducing the datasets. 

\paragraph{Monolingual analysis.} Due to the large availability of data and definition diversity, our experiments are conducted on English datasets. As Multilingual HateCheck \citep{rottger-etal-2022-multilingual} is a multi-lingual extension of HateCheck, in the future, those who have knowledge of other languages and can understand relevant nuances can extend this research to other languages if more datasets with diverse definitions are available. 

\paragraph{Model usage.} Our results are primarily showcased using two models: \texttt{fBERT} and \texttt{RoBERTa}. However, it is not our intention to make generic statements about model generalizability or compare performance between models. To reduce the chances of observations being due to model-specific behavior, we chose two competitive models that are widely used for hate speech detection and revealed different strengths in prior work. Further, the reasonable to high macro-F1 scores indicate that the models are learning the dataset, ruling out a sub-optimal training pipeline. Our results indicate similar performance between \texttt{fBERT} and \texttt{RoBERTa} but it is not guaranteed that this trend extends to other model types. 

We also specifically do not experiment with autoregressive large language models in combination with in-context learning as we do not always know what the training data is. Not only does this lead to complexities due to data leakage but in-context learning is not well-understood and thus cannot be relied upon for a safety-critical setting like hate speech detection.

\paragraph{Taking law into account.} Our analysis does not take the law into account. Where HateCheck gets its target groups from UK law, MHSC looks at the legal definitions from the US. However, the other datasets do not discuss the impact of law on their definition or type of hate speech they are addressing. Thus, we leave it out of our discussion as this would be implicitly encapsulated by the definition (or annotation guidelines) itself. However, the flexibility of our framework can easily allow for legal interpretations to be taken into account, given that a diagnostic dataset includes this type of information and can be modular in its labels to facilitate different laws. 

\paragraph{Geographic bias.} Based on the used datasets, we take a Western perspective for our choice of dominant and non-dominant groups. We include \textit{men} and \textit{white people} as dominant groups as these are the two most popular dominant groups along the identifier categories of gender and race. It is a very subjective take to include dominant groups as a target of hate speech, and our framework leaves it to the user to determine their stance. For instance, HateXplain explicitly considers both \textit{men} and \textit{white people} as targets of hate speech while \textit{DGHS} explicitly excludes them. In no way do we propose or endorse any of the used definitions. When excluded, it becomes even more important to test for these groups to verify and ensure that a model does not accidentally recognize hate speech. This also counts for other aspects from HSC.

\section*{Ethics Statement}
We believe our proposed methodology contributes to the safer deployment of NLP systems. Knowing what is in the data and what the model is capable of is essential for responsible usage. Our proposed methodology intends to verify if a model behaves as originally intended for safety-critical applications. We focus on the task of hate speech detection with an analysis of existing datasets. We show the potential risks of creating or using a dataset without rigorous and fine-grained analysis of dataset creation and model behavior evaluation. Our methodology serves as a tool to identify issues in the dataset if model behavior is not according to expectations thus building toward better-constructed datasets and lesser unexpected inappropriate model performance. 

A note of caution, our methodology does not cover all test items needed when deploying a model for safe and responsible usage. Every project needs thorough testing rounds thus adding project-appropriate test cases too. For instance, edge cases or cases that introduce subjectivity are not covered, which is necessary to understand the full picture of model behavior. As such, this is just a first step. 

\section*{Acknowledgments}
We thank Ilia Markov for his useful comments on the draft version of this camera-ready and Pia Sommerauer for her valuable feedback on an earlier draft. We also thank the anonymous reviewers for their comments that helped improve this paper. All remaining errors are our own. This research was (partially) funded by the Hybrid Intelligence Center, a 10-year programme funded by the Dutch Ministry of Education, Culture and Science through the Netherlands Organisation for Scientific Research. The figures and affiliation emojis have been designed using resources from \textit{flaticon.com}; the crab is created by \textit{mihimihi}, the tulips by \textit{SBTS2018}, analysis chart by \textit{Eucalyp}, neural network by \textit{flatart\_icons}, and the rest of the emojis by \textit{freepik}.

\bibliography{custom}

\appendix

\section{Technical Setup}
\label{app:technical-setup}
We finetune our models with fBERT \citep{sarkar-etal-2021-fbert-neural} ($110$M parameters)\footnote{109.483.778 parameters to be exact}, and RoBERTa-base \citep{liu2019roberta} ($125$M parameters). We follow the pre-processing steps as mentioned in \citet{bourgeade-etal-2023-learn} and replace usernames, URLs, and links with placeholders. For fBERT, we use the original hyperparameters and the learning rate from \citet{bourgeade-etal-2023-learn} as it gave more stable results. We train with a batch size of $8$, warmup on $10\%$ of the total training steps, a weight decay of $0$, and a linear learning rate of $5e^{-5}$ for $3$ epochs. For RoBERTa-base we use the original hyperparameters used for GLUE tasks, sometimes using a batch size of $8$ with gradient accumulation step $2$ due to memory constraints. Warmup is done on $6\%$ of the total training steps, weight decay of $0.1$, and a linear learning rate of $5e^{-5}$ for $10$ epochs. For each dataset, we train $5$ random seeds. Our models are all trained on an RTX 2080 Ti. Each experiment took a maximum of $2$ hours, meaning that the experiments took around $90$ hours in total to finish.  

\paragraph{Model selection} is based on the macro F1 on the validation set. In some cases, the validation loss is much higher ($> 0.1$) than a similar (but slightly less) performing checkpoint. We then choose the one with better loss but slightly worse macro F1.

\paragraph{Splits.} Not all datasets have pre-defined splits, thus in those cases, we create our own random splits with $80\%$ train samples, $10\%$ validation, and $10\%$ test.

\paragraph{Packages.} We use HuggingFace Transformers \citep{wolf-etal-2020-transformers}  and Datasets \citep{lhoest-etal-2021-datasets} for our experiments. 

\section{Decomposition Process}
\label{app:process-reverse-engineering}
For each dataset, we introduce it, decompose the definition (and annotation guidelines, if provided) using Hate Speech Criteria \textbf{(HSC)}, and provide the decomposed definition. We conclude with our model behavior expectations. These datasets are chosen based on their wide usage and diversity in their hate speech definition. We mark the different aspects in the definitions in the following way: \underline{target group}, [dominance], \texttt{consequence}, and \{explicit reference\}.

\paragraph{The \textbf{\textsc{TalatHovy} \citep{waseem2016hatefulTalat}} dataset} focuses particularly on \textit{racism} and \textit{sexism} toward minorities. It consists of $16914$ tweets, of which $20\%$ are \textit{sexist}, $11.66\%$ are \textit{racist}, and $68.34\%$ are neither. 

\begin{tcolorbox}[title=\textsc{TalatHovy} Definition, mybox]
   (Page 2, Section Data) ``A tweet is offensive if it
    \begin{enumerate}
        \item uses a sexist or racial slur
        \item attacks a minority
        \item seeks to silence a minority
        \item criticizes a minority (without a well founded argument)
        \item promotes, but does not directly use, hate speech or violent crime
        \item criticizes a minority and uses a straw man argument
        \item blatantly misrepresents truth or seeks to distort views on a              minority with unfounded claims
        \item shows support of problematic hash tags. E.g. “\#BanIslam”, “\#whoriental”, “\#whitegenocide”
        \item negatively stereotypes a minority
        \item defends xenophobia or sexism
        \item contains a screen name that is offensive, as per the previous criteria, the tweet is ambiguous (at best), and the tweet is on a topic that satisfies any of the above criteria.''       
    \end{enumerate}
\end{tcolorbox}

\noindent \textbf{Decomposed:} For this dataset, hate speech is defined as language targeted at a [non-dominant person or group] based on their \underline{race and gender} and \texttt{incites violence and hate or insults a group} through the usage of \{{negative stereotypes, group characteristics, or slur\}. \textbf{Expectations:} We expect models to perform well for non-dominant groups based on their race and gender, as well as religion and nationality due to their closeness to racism in the dataset. Furthermore, it should recognize incitement of violence, hate, and group insult. It should also capture all three types of explicit references to a group. As the dataset focuses on race and gender-based minorities, we do not expect the model to classify dominant groups from the same category (i.e.\ men and white people) or other target groups as hate speech. As incitement of discrimination is not explicitly mentioned, it is unknown if the model can capture this phenomenon and generalize well. 

\paragraph{The \textbf{\textsc{Davidson} \citep{davidson2017automated}} dataset} targets hate speech in general. The dataset comprises $24802$ tweets, of which $5.77\%$\footnote{The percentages were calculated from the dataset from \url{https://github.com/t-davidson/hate-speech-and-offensive-language}. These values might slightly deviate from the ones mentioned in the original paper.} are \textit{hate speech}, $77.43\%$ \textit{offensive}, and $16.80\%$ neither. 

\begin{tcolorbox}[title=\textsc{Davidson} Definition, mybox]
    (Page 1, Introduction) ``We define hate speech as language that is used to expresses hatred towards a targeted group or is intended to be derogatory, to humiliate, or to insult the members of the group. In extreme cases this may also be language that threatens or incites violence, but limiting our definition only to such cases would exclude a large proportion of hate speech. Importantly, our definition does not include all instances of offensive language because people often use terms that are highly offensive to certain groups but in a qualitatively different manner. For example some African Americans often use the term n*gga in everyday language online \citep{warner-hirschberg-2012-detecting}, people use terms like h*e and b*tch when quoting rap lyrics, and teenagers use homophobic slurs like f*g as they play video games. Such language is prevalent on social media \citep{wang2014cursing}, making this boundary condition crucial for any usable hate speech detection system'' \\ 

    (Page 2, Section Data) ``They [Annotators] were provided with our definition along with a paragraph explaining it in further detail. Users were asked to think not just about the words appearing in a given tweet but about the context in which they were used. They were instructed that the presence of a particular word, however offensive, did not necessarily indicate a tweet is hate speech.''
\end{tcolorbox}

\noindent \textbf{Decomposed:} For this dataset, hate speech is defined as language targeted at a \underline{group} to \texttt{incite hate} or a \underline{person of a group} to \texttt{insult or incite violence} through the usage of \{group characteristics\}. \textbf{Expectations:}  No target group or anything about dominance is mentioned. Thus, it is unclear on which target types the model will work well, nor if dominant groups are also covered. The same holds for \textit{slurs} and \textit{stereotypes}. However, we do expect the models to capture incitement of hate, violence, and group insult, except for incitement of discrimination, which is unknown. 

\paragraph{The \textbf{\textsc{Measuring Hate Speech Corpus} \citep{kennedy2020constructing}}  dataset} targets a variety of hate speech types and contains $39565$ comments. Of these comments, $26.2\%$ is hate speech and $73.8\%$ is not.\footnote{The instances come with a score which we translate to labels (i.e.\ hate speech or not) using the threshold from the original paper: \url{https://huggingface.co/datasets/ucberkeley-dlab/measuring-hate-speech}}

\begin{tcolorbox}[title=\textsc{MHSC} Definition, mybox]
   (Page 7, Section 3.1) ``We draw from the legal definition of hate crimes in the United States that protects against discriminatory actions targeting one of the following protected groups: race, religion, ethnicity, nationality, gender, sexual orientation, gender identity, and disability. In identifying groups within these broad categories, we include subjugated groups that have been discriminated against in the United States, as well as power-dominant groups who have not. Targeting of a group or an individual on the basis of their membership in a group is common to most definitions of hate speech \citep{sellars2016defining}. Not only do we adopt this convention, but we allow for intersectional or overlapping identities to be selected for further analysis. We consider intersectional identities and the possibility of compounding hate speech directed at an individual who belongs to multiple groups. \\
   
   Speech can also lead to individual acts of violence and when targeted against a group, genocide and extermination. The “dangerous speech” framework ties the effects of hateful speech to actions that it can incite (Benesch et al. 2018). Dehumanization, such as radio broadcasts in Rwanda referring to the Tutsi people as cockroaches, is directly linked to later genocidal killing of that group. Incitement towards violence is a narrowly defined concept under US law, and the dangerous speech framework that we use takes a broader view of the link between cause and effect. \citet{sellars2016defining} points out that the accumulated affects of anti-Semitic or racist speech can have multi-generational impacts on the well-being of individuals in a group born long after hateful speech was original created. Given the complexities of these concepts, we focus on calls to individual violence or collective extermination, with the idea that these are the final step after expression of hate and deeming a group inferior or inhuman. \\
   
   Table 2 describes the eight levels of our theorized hate speech scale. The positive levels on the scale designate hate speech of increasing severity. Unlike many existing scales, our typology includes both neutral and positive identity speech, represented by 0 and negative values, respectively. Following Anti-Defamation League (2016) and Stanton (2013), we place speech supporting the systematic killing of a specific group as the most severe form of hate speech. Viewing other types of hate speech as pathways to genocide, we pay special attention to individuals threats of violence and dehumanization that may justify violence.''
\end{tcolorbox}

\noindent \textbf{Decomposed:} For this dataset, hate speech is defined as language targeted at a \underline{person or group based on their} \underline{race}, \underline{religion}, \underline{ethnicity}, \underline{nationality}, \underline{gender}, \underline{sexual orientation}, \underline{gender identity}, \underline{age}, and \underline{disability}. \texttt{It incites hate, violence, and discrimination} for both [dominant and (non-)dominant groups} through the usage of \{group characteristics\}. \textbf{Expectations:} Recognizes hate speech against all target types, including dominant groups. While we expect good generalization for incitement of hate, violence, and discrimination, it is unclear if group insult will work well. Slurs and stereotypes should be captured.

\paragraph{The \textbf{\textsc{Dynamically Generated Hate Speech}} \citep{vidgen2021learning} dataset} is dynamically created with human-in-the-loop. The annotators are provided with extensive annotator guidelines with many annotation rounds. It consists of 41255 examples, of which $54.0\%$ is hate speech and $46.0\%$ is not.\footnote{Note that this is the most balanced dataset out of all.} 

\begin{tcolorbox}[title=\textsc{DGHS} Definition, mybox]
    (Pages 3-4, Section 3; Annotation Guidelines\footnote{\url{https://github.com/bvidgen/Dynamically-Generated-Hate-Speech-Dataset/blob/main/Dynamically\%20Generated\%20Hate\%20Dataset\%20-\%20annotation\%20guidelines.pdf}}) ``‘Hate’ is defined as “abusive speech targeting specific group characteristics,
such as ethnic origin, religion, gender, or sexual
orientation.” \citep{warner-hirschberg-2012-detecting} \\

\textbf{3.1 Types of hate:} \\
\indent \textit{Derogation}: Content which explicitly attacks, demonizes, demeans or insults a group. This resembles similar definitions from \citet{davidson2017automated}, who define hate as content that is ‘derogatory’, \citet{waseem2016hatefulTalat} who include ‘attacks’ in their definition, and \citet{zampieri-etal-2019-predicting} who include ‘insults’. \\
\indent \textit{Animosity}: Content which expresses abuse against a group in an implicit or subtle manner. It is similar to the ‘implicit’ and ‘covert’ categories used in other taxonomies (Waseem et al., 2017; \citet{vidgen2020detecting}; \citet{kumar-etal-2018-benchmarking}). \\ 
\indent \textit{Threatening language}: Content which expresses intention to, support for, or encourages inflicting harm on a group, or identified members of the group. This category is used in datasets by \citet{hammer2014detecting}, \citet{golbeck2017large} and \citet{anzovino2018automatic}. Support for hateful entities Content which explicitly glorifies, justifies or supports hateful actions, events, organizations, tropes and individuals (collectively, ‘entities’). \\
\indent \textit{Dehumanization}: Content which ‘perceiv[es] or
treat[s] people as less than human’ \citep{haslam2016recent}. It often involves describing groups as leeches, cockroaches, insects, germs or rats \citep{mendelsohn2020framework}. \\

\textbf{3.2 Targets of hate:} Hate can be targeted against any vulnerable, marginalized or discriminated-against group. We provided annotators with a non-exhaustive list of 29 identities to focus on (e.g., women, black people, Muslims, Jewish people and gay people), as well as a small number of intersectional variations (e.g., ‘Muslim women’). They are given in Appendix A. Some identities were considered out-of-scope for Hate, including men, white people, and heterosexuals.''
\end{tcolorbox}

\noindent \textbf{Decomposed:} 
For this dataset, hate speech is defined as language targeted at a \underline{person or group based on} \underline{religion}, \underline{race}, \underline{gender}, \underline{sexual orientation}, \underline{nationality}, and \underline{disability} and \texttt{insults the group or incites hate or violence} for [especially non-dominant groups] through the usage of \{group characteristics, slurs, and stereotypes\}. \textbf{Expectations:} We expect good performance on all the target types in HateCheck. Dominant groups will not be seen as part of hate speech. Except for incitement of discrimination, which we do not know will be captured or not, group insult and incitement of hate and violence should have good performance. All types of explicit references should get good performance.

\paragraph{The \textbf{\textsc{HateXplain} \citep{mathew2021hatexplain}} dataset} comprises 20148 examples and has classes for offensive language and undecided instances. The class distribution is $29.5\%$ hateful, $27.2\%$ offensive, $38.8\%$ normal, and $4.5\%$ undecided. 

 \begin{table}[h]
        \centering
        \scalebox{0.7}{
        \begin{tabular}{ll}
            \toprule
            \textbf{Target Groups} & \textbf{Categories} \\ \midrule
            Race & African, Arabs, Asians, Caucasian, Hispanic \\
            Religion & Buddhism, Christian, Hindu, Islam, Jewish \\
            Gender & Men, Women \\
            Sexual Orientation & Heterosexual, Gay \\
            Miscellaneous & Indigenous, Refugee/Immigrant, None, Others \\
            \bottomrule
        \end{tabular}}
        \caption{(Page 3, Table 3) Target groups considered in \textsc{HateXplain}}
        \label{tab:tg-hx}
    \end{table}

\noindent \textbf{Decomposed:} For this dataset, hate speech is defined as language targeted at a \underline{person or group based on their} \underline{race}, \underline{religion}, \underline{gender}, or \underline{sexual orientation} from both [dominant and non-dominant groups] through the usage of \{group characteristics\}. \textbf{Expectations:} We expect good performance on \textit{race, religion, gender, nationality}, and \textit{sexual orientation}, also for dominant groups. Other target types and aspects are unknown since they are unmentioned.

\paragraph{The \textbf{\textsc{Founta} \citep{founta2018large}} dataset} consists of $4.97\%$ hateful, $27.15\%$ abusive, $14.03\%$ spam, and $53.85\%$ normal tweets. 

\begin{tcolorbox}[title=\textsc{Founta} Definition, mybox]
    (Page 5, Section Step 2: Exploratory Rounds) ``\textbf{Hate Speech:} Language used to express hatred towards a targeted individual or group, or is intended to be derogatory, to humiliate, or to insult the members of the group, on the basis of attributes such as race, religion, ethnic origin, sexual orientation, disability, or gender. \citep{davidson2017automated}, \citep{badjatiya2017deep}, \citep{warner-hirschberg-2012-detecting}, \citep{schmidt-wiegand-2017-survey}, \citep{djuric2015hate}.''
    
\end{tcolorbox}

\noindent \textbf{Decomposed:} For this dataset, hate speech is language targeted at a \underline{person or group based on their} \underline{race}, \underline{religion}, \underline{ethnic origin}, \underline{sexual orientation}, \underline{disability}, or \underline{gender etc.} and \texttt{insults a group or incites hate} through the usage of \{group characteristics\}. \textbf{Expectations:} We expect good performance on all target types. We also expect the model to capture incitement of hate and group insult. The performance on dominance and the rest of the aspects is unknown.

\section{Results on Individual Datasets}
\label{app:results-indiv-datasets}

\begin{figure*}[h!]
    \centering
    \subfloat[fBERT.]{
        \includegraphics[width=0.45\textwidth]{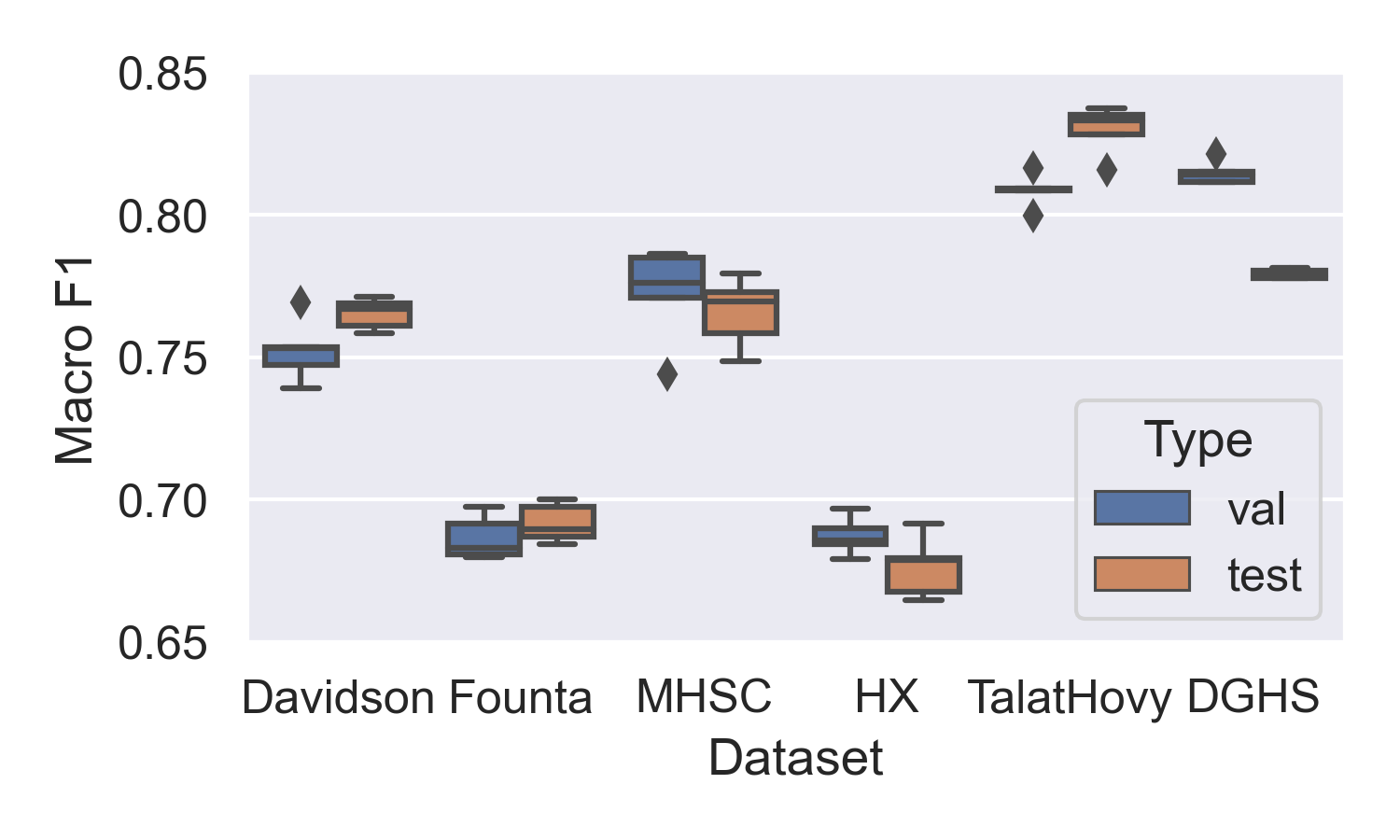}
        \label{fig:indiv-fbert}
    } 
    \subfloat[RoBERTa.]{
        \includegraphics[width=0.45\textwidth]{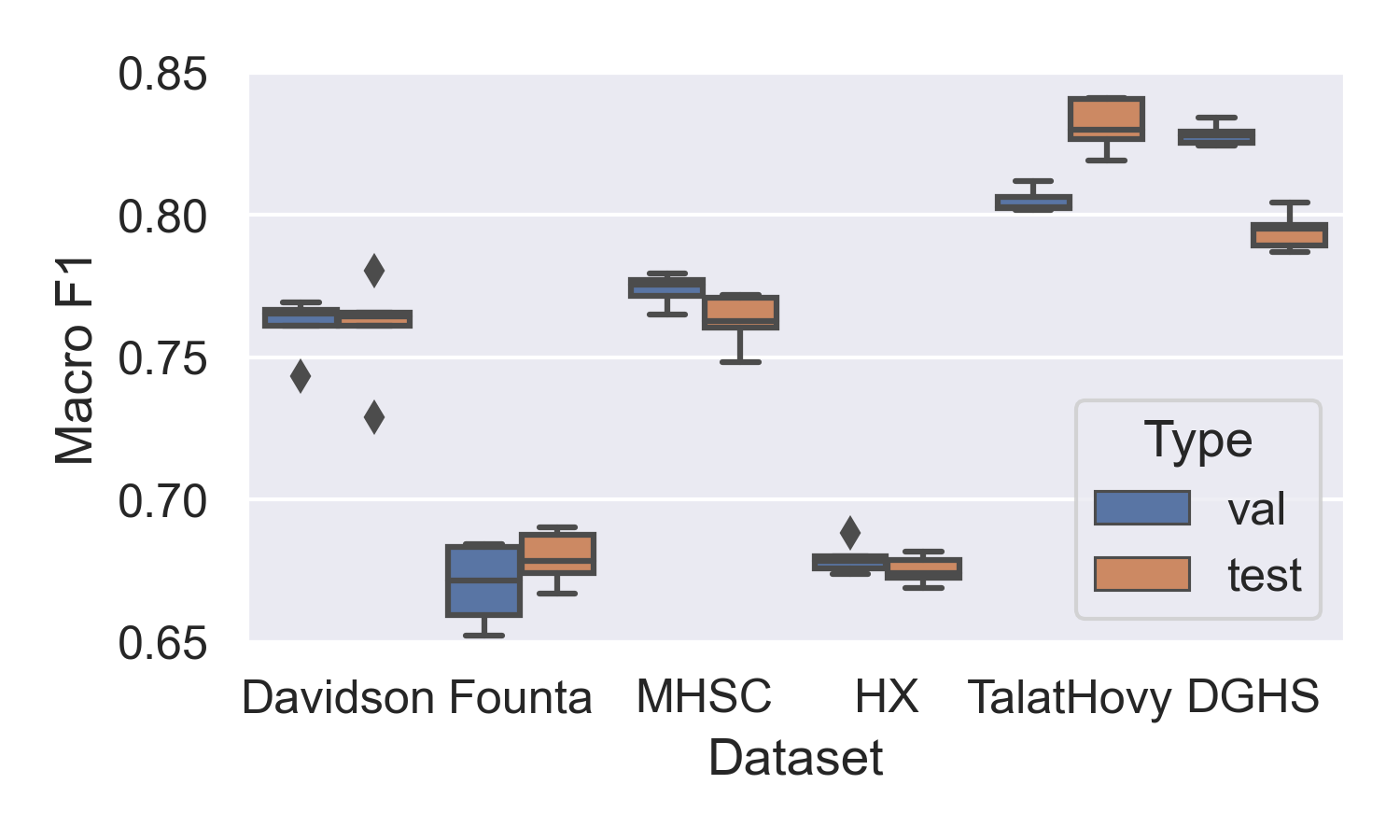}
        \label{fig:indiv-roberta}
    } 
    \caption{Macro F1 scores on the respective validation and test set for all the \textit{six} datasets.}
    \label{fig:intra_dataset}
\end{figure*}

In this Section, we show the macro F1 obtained on the respective validation and test sets of each dataset in Figure~\ref{fig:intra_dataset}, for both fBERT and RoBERTa. 

We observe that most datasets achieve average to good performance on their respective validation and test sets (Figure~\ref{fig:intra_dataset}). Both fBERT (Figure~\ref{fig:indiv-fbert}) and RoBERTa (Figure~\ref{fig:indiv-fbert}) appear to yield similar results. Particularly, we see for \textbf{TalatHovy} and \textbf{DGHS} a high Macro F1, in the range of $0.8$ to $0.825$ with low variation in seeds, except for \textbf{DGHS} where one seed underperforms ($0.35$).\footnote{For clarity in the plot, we limit the y-axis.} The \textbf{MHSC} dataset also achieves a higher Macro F1 ($\sim 0.775$) but with slightly more variation across seeds. The rest of the datasets also have slightly more variation across seeds but lesser performance in comparison, yielding decent results ($\sim 0.675 - 0.75$).  

\section{Analysis of DGHS' Composition}
\label{app:vidgen-datacomp}
To identify the influence of DGHS's dataset composition on its superior performance, we experiment with augmenting the training set size. In Table~\ref{tab:vidgen-sizess} we show the label distribution of the two different \textit{DGHS} training sets we experiment with, using fBERT. \textbf{DGHS$_{Small}$}: a smaller training set size that approximates the size of the smallest dataset, the TalatHovy dataset. \textbf{DGHS$_{Imbalanced}$}: the imbalanced dataset, where we keep 10\% of the instances hate and the rest of the 90\% not hate, essentially also decreasing the size. This corresponds to many of the datasets where there are very few \textit{hate speech} instances but many \textit{non-hate speech} ones. 

\begin{table}[h!]
    \centering
    \begin{tabular}{ccc}
        \toprule
         & \textbf{\#Hate} & \textbf{\#Not Hate} \\
        \midrule
        DGHS$_{Original}$ & 17740 & 15184 \\
        DGHS$_{Small}$ & 7537 & 6463 \\
        DGHS$_{Imbalanced}$ & 1687 & 15184 \\
        \bottomrule
    \end{tabular}
    \caption{Data Summary}
    \label{tab:vidgen-sizess}
\end{table}

\begin{figure}[h!]
    \centering
    \subfloat[DGHS's Test Set.]{
        \includegraphics[width=0.45\textwidth]{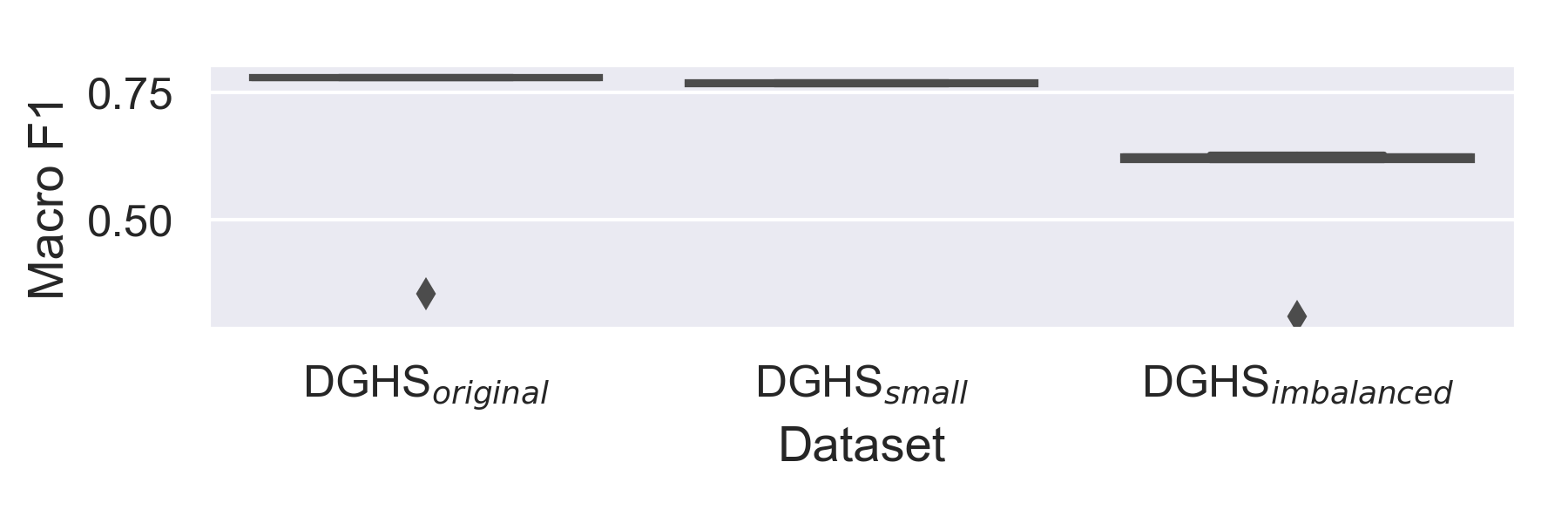}
        \label{fig:test-v}
    } \\
    \subfloat[HateCheck.]{
        \includegraphics[width=0.45\textwidth]{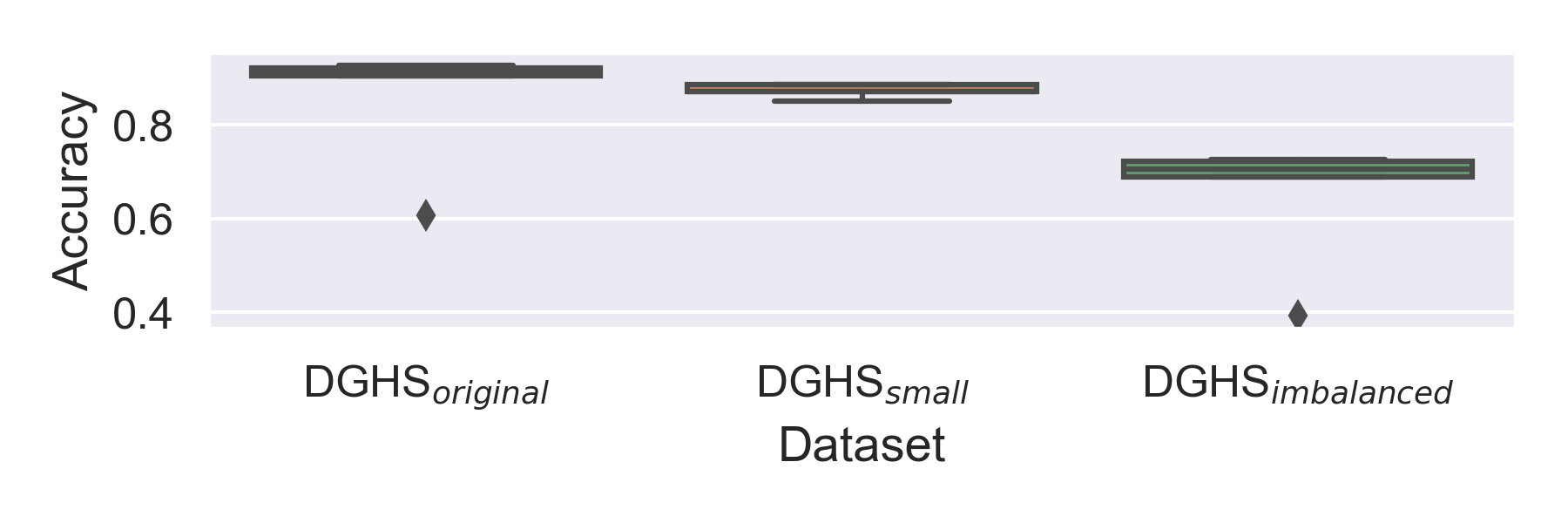}
        \label{fig:hc-v}
    } \\
    \subfloat[Cross-evaluation.]{
        \includegraphics[width=0.45\textwidth]{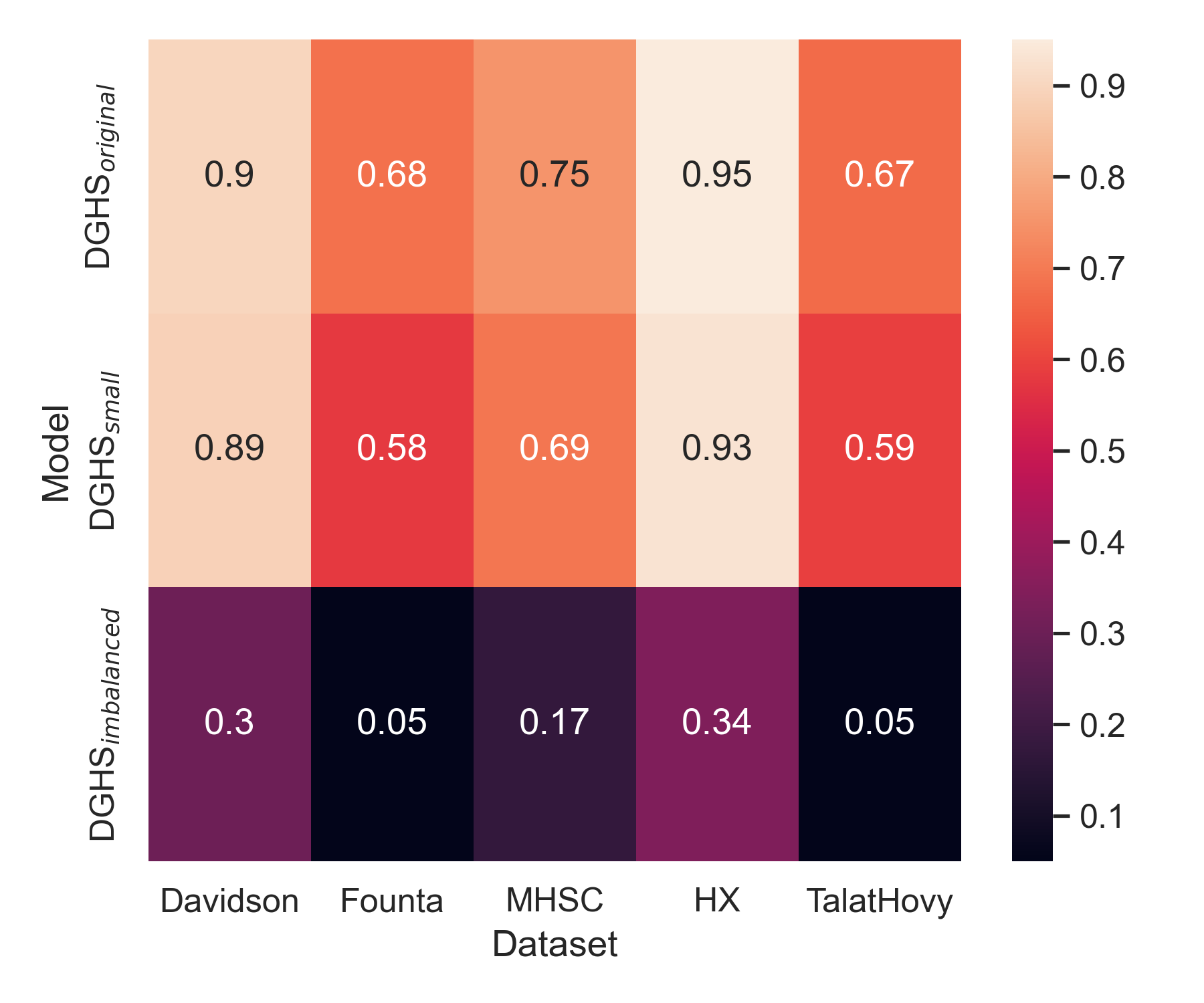}
        \label{fig:cross-eval-v}
    } 
    \caption{Results of testing different data compositions for DGHS's dataset.}
    \label{fig:v-performance}
\end{figure}

The results on the dataset's respective test set, overall HateCheck, and cross-evaluation can be found in Figure \ref{fig:v-performance}. Sizing down the training set with DGHS$_{Small}$ does not have as much impact on the results overall. The results are slightly lower than the results with the original training set. However, results of DGHS$_{Imbalanced}$ clearly take a hit when cross-evaluating, indicating that it is essential to have a large amount of \textit{hate speech} samples for good cross-evaluation and performance in general.

\end{document}